\documentclass[10pt,twocolumn,letterpaper]{article}

\usepackage{cvpr}
\usepackage{times}
\usepackage{epsfig}
\usepackage{graphicx}
\usepackage{amsmath}
\usepackage{amssymb}

\usepackage{subfigure}
\usepackage[section]{algorithm}
\usepackage{algorithmic}

\usepackage{makecell}
\usepackage{multirow}
\usepackage{array}
\usepackage{colortbl}

\usepackage{pdfpages}
\newcommand{\cD}{\mathcal{D}}

\newcommand{\cI}{\mathcal{I}}

\newcommand{\cL}{\mathcal{L}}

\newcommand{\R}{\mathbb{R}}

\newcommand{\suml}[2]{\sum\limits_{#1}^{#2}}

\newcolumntype{L}[1]{>{\raggedright\let\newline\\\arraybackslash\hspace{0pt}}m{#1}}
\newcolumntype{C}[1]{>{\centering\let\newline\\\arraybackslash\hspace{0pt}}m{#1}}
\newcolumntype{R}[1]{>{\raggedleft\let\newline\\\arraybackslash\hspace{0pt}}m{#1}}

\definecolor{grayB}{rgb}{0.85,0.85,0.85}
\usepackage[pagebackref=true,breaklinks=true,letterpaper=true,colorlinks,bookmarks=false]{hyperref}

\cvprfinalcopy 

\def\cvprPaperID{306} 

\ifcvprfinal\fi
\begin{document}

\title{On learning optimized reaction diffusion processes for effective image restoration}

\author{Yunjin Chen$^{1,2}$\quad \quad \quad \quad Wei Yu$^{1}$ \quad \quad \quad \quad Thomas Pock$^{1,3}$\\
$^1$Graz University of Technology \quad $^2$National University of Defense Technology \\ 
\quad $^3$Digital Safety \& Security Department, AIT Austrian Institute of Technology GmbH\\
}

\maketitle

\begin{abstract}
  For several decades, image restoration remains an active research
  topic in low-level computer vision and hence new approaches are
  constantly emerging. However, many recently proposed algorithms
  achieve state-of-the-art performance only at the expense of very
  high computation time, which clearly limits their practical
  relevance. In this work, we propose a simple but effective approach with both
  high computational efficiency and high restoration quality. We
  extend conventional nonlinear reaction diffusion models by several
  parametrized linear filters as well as several parametrized
  influence functions.  We propose to train the parameters of the
  filters and the influence functions through a loss based
  approach. Experiments show that our trained nonlinear reaction
  diffusion models largely benefit from the training of the parameters
  and finally lead to the best reported performance on common test
  datasets for image restoration. Due to their structural simplicity,
  our trained models are highly efficient and are also well-suited for
  parallel computation on GPUs.
\end{abstract}

\vspace*{-0.5cm}
\section{Introduction}
Image restoration is the process of estimating uncorrupted images from
noisy or blurred ones.  It is one of the most fundamental operation in
image processing, video processing, and low-level computer vision.
There exists a huge amount of literature addressing the topic of image
restoration problems, see for example \cite{milanfar2013tour} for a
survey.  Broadly speaking, most state-of-the-art techniques mainly
concentrate on achieving utmost image restoration quality, with little
consideration on the computational efficiency \cite{EPLL, LSSC, WNNM}.
However, there are two notable exceptions, BM3D \cite{BM3D} and the
recently proposed Cascade of Shrinkage Fields (CSF) \cite{CSF2014}
model, which simultaneously offer high efficiency and high image
restoration quality.

\begin{figure}[t!]
\vspace*{-0.2cm}
\centering
\subfigure[Truncated convex]{\includegraphics[width=0.235\textwidth]{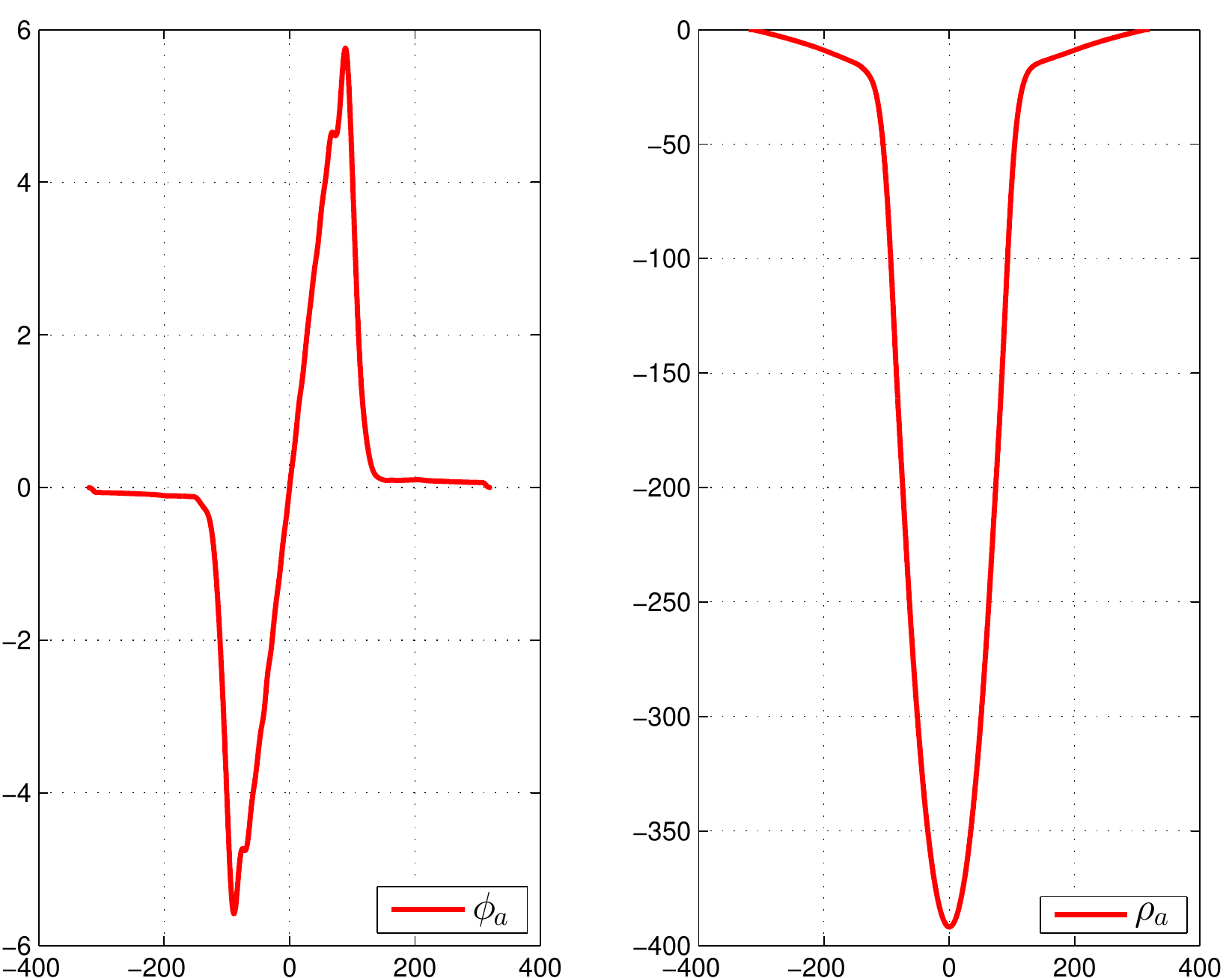}}\hfill
\subfigure[Negative Mexican hat]{\includegraphics[width=0.235\textwidth]{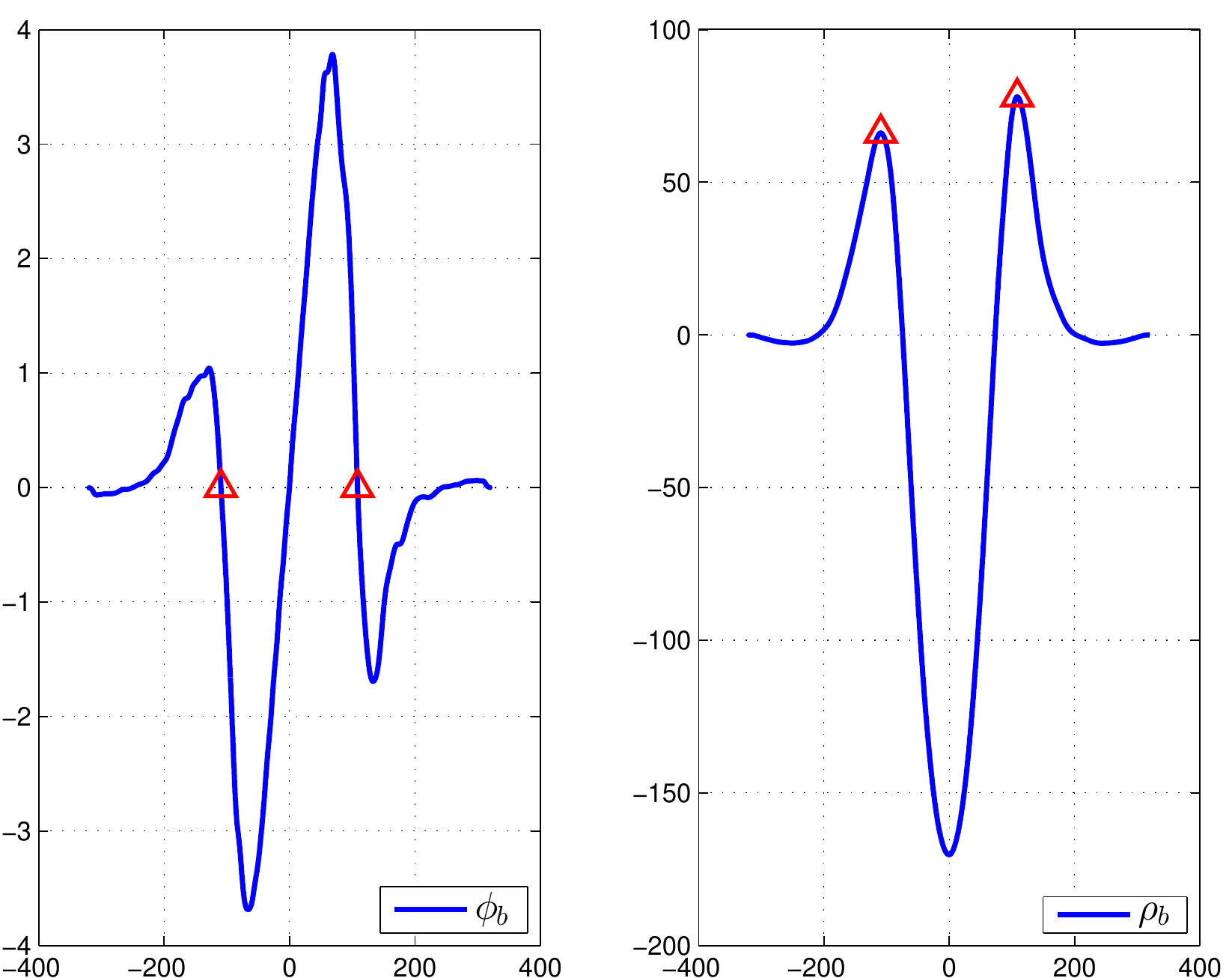}}\\
\vspace*{-0.3cm}
\subfigure[Truncated concave]{\includegraphics[width=0.235\textwidth]{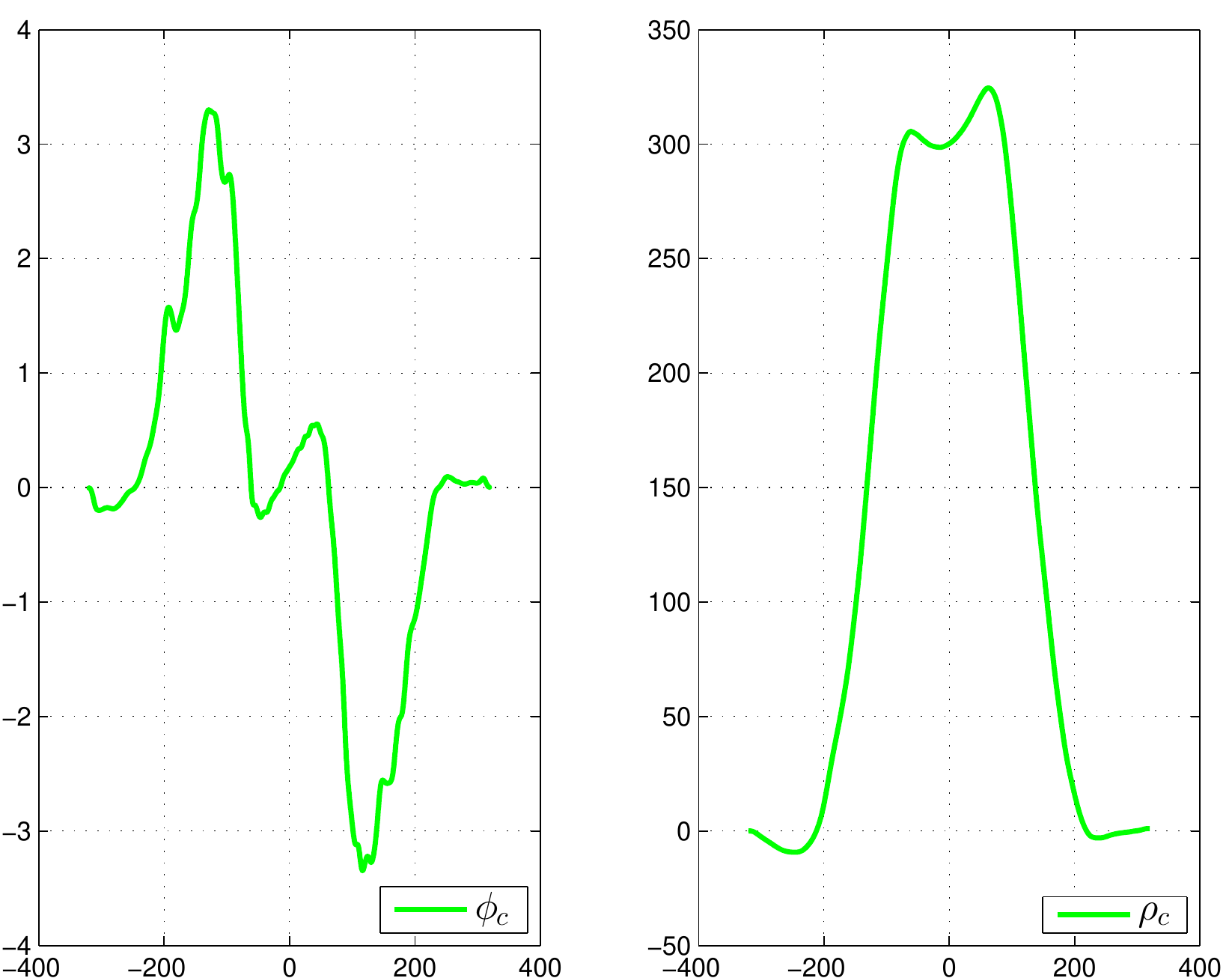}}\hfill
\subfigure[Double-well penalty]{\includegraphics[width=0.235\textwidth]{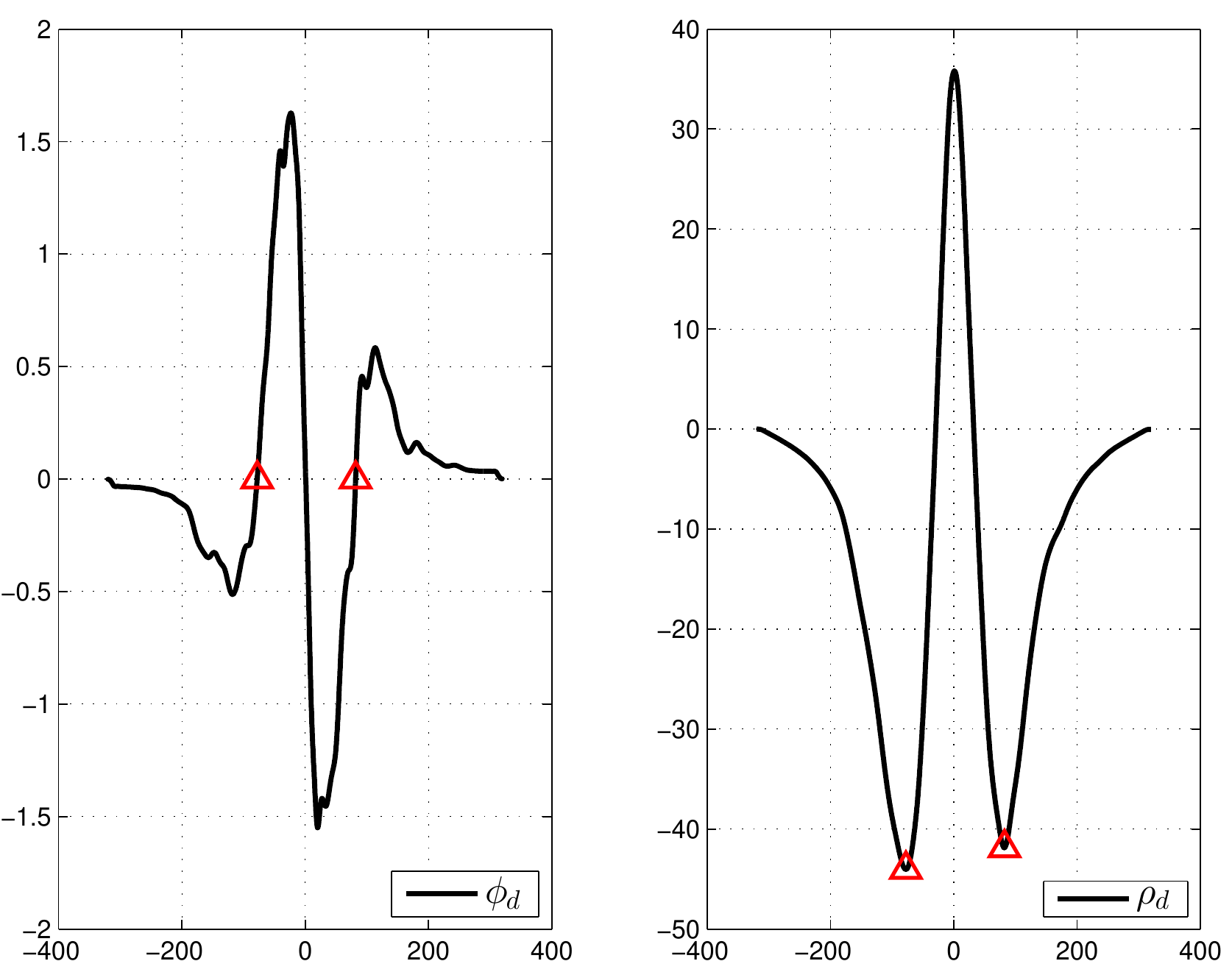}}
\caption{The figure shows four characteristic influence functions
  (left plot in each subfigure) together with their corresponding
  penalty functions (right plot in each subfigure), learned by our
  proposed method. A major finding in this paper is that our learned
  penalty functions significantly differ from the usual penalty
  functions adopted in partial differential equations and energy minimization
  methods. In contrast to their usual robust smoothing properties
  which is caused by a single minimum around zero, most of our learned
  functions have multiple minima different from zero and hence are
  able to enhance certain image structures. See Sec. \ref{sec:denoising} for more
  information.}\label{functions}\vspace{-0.5cm}
\end{figure}

It is well-known that BM3D is a highly engineered method, specialized
for Gaussian noise. Moreover, it involves a block matching process,
which is challenging for parallel computation on GPUs, alluding to the
fact that it is not straightforward to accelerate BM3D algorithm on
parallel architectures. In contrast, the recently proposed CSF model
offers high levels of parallelism, making it well suited for GPU
implementation, thus owning high computational efficiency.

Among the approaches to tackle the problem of image restoration,
nonlinear anisotropic diffusion \cite{PMmodel, anisotropicbook} defines a class of
efficient approaches, as each diffusion step merely contains the
convolution operation with a few linear filters. A nonlinear diffusion process usually corresponds to certain 
Partial Differential Equation (PDE) formulation. However, up to now,
the image restoration quality of diffusion based approaches is still
far away from the state-of-the-art, although with many improvements
\cite{hajiaboli2011anisotropic, didas2009properties,
  plonka2008nonlinear, guidotti2011two}.

We give a brief review of nonlinear diffusion based approaches and then introduce our proposed diffusion model. 
In the seminal work \cite{PMmodel}, Perona and Malik (P-M) demonstrated that nonlinear diffusion models 
yield very impressive results for image processing. This has given rise to 
many revised models with various formulations. A notable variant is the so-called biased anisotropic diffusion 
(also known as reaction diffusion) proposed by Nordstr{\"o}m \cite{biased}, which introduces a bias term (forcing term) to 
free the user from the difficulty of specifying an appropriate stopping time for the P-M diffusion process. 
This additional term reacts against the strict smoothing effect of the pure P-M diffusion, therefore resulting in a nontrivial steady-state. 

Tsiotsios \etal \cite{tsiotsios2013choice} discussed the choice of some crucial parameters in the P-M model, such as 
the diffusivity function, the gradient threshold parameter and the stopping time of the iterative process. 
Some works consider modification to the diffusion term or the reaction term for the reaction diffusion model 
\cite{esclarin1997image, cottet1993image, biased, plonka2008nonlinear, acton2001oriented}, \eg, 
Acton \etal \cite{acton2001oriented} and Plonkna \etal \cite{plonka2008nonlinear} exploited 
a more complicated reaction term to enhance oriented textures; \cite{surya2014weighted, barcelos2003well} proposed 
to replace the ordinary diffusion term with a flow equation based on mean curvature. 
Gilboa \etal \cite{gilboa2002forward} proposed a forward and backward diffusion process, which incorporates explicit 
inverse diffusion with negative diffusivity coefficient by carefully choosing the diffusivity function. The resultant diffusion 
processes can adaptively switch between forward and backward diffusion process. In a latter work \cite{welk2009theoretical}, 
the theoretical foundations for discrete forward-and-backward diffusion filtering were investigated. 
Researchers also propose to exploit higher-order nonlinear diffusion filtering, which involves larger linear filters, 
\eg, fourth-order diffusion models \cite{hajiaboli2011anisotropic, didas2009properties, guidotti2011two}.
Meanwhile, theoretical properties about the stability and local feature enhancement of higher-order nonlinear diffusion filtering 
are established in \cite{didas2005stability}. 

It should be noted that all the above mentioned diffusion processes are handcrafted models. 
It is a generally difficult task to design a good-performing PDE for a specific image processing problem because 
good insights into this problem and a deep understanding of the behavior of the PDEs are usually required. 
Therefore, some researcher propose to learn PDEs from training data via an optimal control approach \cite{liu2010learning}. 
Unfortunately, at present \cite{liu2010learning} is the sole previous work we can find in this direction. 
The basic idea of our approach is the same as \cite{liu2010learning}, but we go much further and our proposed model 
is much more expressive. 

\subsection{Motivation and Contributions}
In this paper we focus on nonlinear diffusion process due to its high efficiency and propose 
a trainable nonlinear diffusion model, which is parameterized by the linear filters and the influence functions. 
The trained diffusion model contains many special influence functions (see Fig. \ref{functions} for an illustration), which 
greatly differ from usual influence functions employed in conventional diffusion models. It turns out that 
the trained diffusion processes can lead to effective image restoration with state-of-the-art performance, 
while preserve the property of high efficiency of diffusion based approaches. 
At present, we are not aware of any previous works that simultaneously optimize the linear filters and influence functions 
of a nonlinear diffusion process\footnote{
Even though the linear filters and penalty functions in the image prior model \cite{RothFOE2009, ChenPRB13} 
can be trained simultaneously, the 
penalty function is optimized only in the sense that the weight $\alpha$ of certain fixed function 
(\eg, $\alpha \cdot \text{log}(1+z^2)$) can be tuned. Our approach can exploit much more generalized penalty functions 
(actually arbitrary functions), which is intractable in those previous models.}. 

Our proposed nonlinear diffusion process has several remarkable benefits as follows: 
\vspace*{-0.2cm}
\begin{itemize}
\setlength\itemsep{-0.4em}
\item[1)] It is conceptually simple as it is just a time-dynamic
  nonlinear reaction diffusion model with trained filters and influence
  functions;
\item[2)] It has broad applicability to a variety of image restoration
  problems.  In principle, all existing diffusion based models can be
  revisited with appropriate training;
\item[3)] It yields excellent results for several tasks in image
  restoration, including Gaussian image denoising, and JPEG
  deblocking;
\item[4)] It is computationally very efficient and well suited for
  parallel computation on GPUs.
\end{itemize}

\vspace*{-0.4cm}\section{Proposed reaction diffusion process}
We start with conventional nonlinear diffusion processes, 
then propose a training based reaction diffusion model for image restoration. 
Finally we show the relations between the proposed model and existing image restoration models. 

\subsection{Perona and Malik diffusion model}
In the whole paper, we stick to the fully discrete setting, where
images are represented as column vectors, \ie, $u \in
\R^{N}$. Therefore, the discrete version of the well-known
Perona-Malik type nonlinear diffusion process \cite{PMmodel} can be
formulated as the following discrete PDE with an explicit finite difference scheme
\begin{equation}\label{pmmodel}
\vspace*{-0.2cm}
  \hspace*{-0.3cm}\frac{u_{t+1} - u_t}{\Delta t} = -\suml{i = \{x,y\}}{}\nabla_i^\top \Lambda (u_t) \nabla_i u_t
  \doteq -\suml{i = \{x,y\}}{}\nabla_i^\top \phi(\nabla_i u_t) \,,
\end{equation}
where matrices $\nabla_x$ and $\nabla_y \in \R^{N \times N}$ are
finite difference approximation of the gradient operators in
$x$-direction and $y$-direction, respectively and $\Delta t$ denotes
the time step. $\Lambda(u_t) \in \R^{N \times N}$ is defined as a
diagonal matrix
\[
\Lambda(u_t) = \text{diag}\left( g\left(\sqrt{(\nabla_x u_t)^2_p +
      (\nabla_y u_t)^2_p}\right)\right)_{p = 1, \cdots, N}\,,
\]
where function $g$ is known as edge-stopping function
\cite{ROBUSTANISOTROPIC} or diffusivity function
\cite{anisotropicbook}, a typical $g$ function given as $g(z) =
{1}/{(1+z^2)}$.  If ignoring the coupled relation between $\nabla_x u$
and $\nabla_y u$, the P-M model can be also written as the second
formula on the right side in \eqref{pmmodel}, where $\phi(\nabla_i u)
= \left( \phi(\nabla_i u)_1, \cdots, \phi(\nabla_i u)_N \right)^\top
\in \R^N$ with function $\phi(z) = zg(z)$, known as influence function
\cite{ROBUSTANISOTROPIC} or flux function \cite{anisotropicbook}.  In
the upcoming subsection, we will stick to this decoupled formulation.
\subsection{Proposed nonlinear diffusion model}
Clearly, the matrix-vector product, $\nabla_x u$ can be interpreted as a
2D convolution of $u$ with the linear filter $k_x = [-1, 1]$
($\nabla_y$ corresponds to the linear filter $k_y = [-1, 1]^\top$).
Intuitively, in order to improve the capacity of the diffusion model,
we can employ more filters of larger kernel size, in contrast to
previous works that typically involve few filters with relatively
small kernel size. We can additionally consider different influence
functions for different filters, rather than an unique one.  Moreover,
the parameters of each iteration can vary across iterations. Taking the reaction term into account, our proposed
nonlinear reaction diffusion model is formulated as
\vspace*{-0.4cm}\begin{equation}\label{ourreactionmodel}
\frac{u_{t} - u_{t-1}}{\Delta t} = 
-\underbrace{\suml{i = 1}{N_k}{K_i^t}^\top \phi_i^t(K_i^t u_{t-1})}_\text{diffusion term}~ - ~
\underbrace{\psi(u_{t-1}, f_n)}_\text{reaction term},
\vspace*{-0.2cm}
\end{equation}
where $K_i \in \R^{N \times N}$ is a highly sparse matrix, implemented as 2D convolution of the image $u$ 
with the filter kernel $k_i$, i.e., $K_i u \Leftrightarrow k_i*u$, $K_i$ is a set of linear filters and $N_k$ is the number of filters. 
Function $\phi_i$ operates point-wise to the filter response $K_i u$. 
In practice, we set $\Delta t = 1$, as we can freely scale the formula on the right side. 
Note that in our proposed diffusion model, the influence functions are adjustable and can be different from each other. 

The specific formulation for the reaction term $\psi(u)$ depends on
applications.  For classical image restoration problems, such as
Gaussian denoising, image deblurring, image super resolution and image
inpainting, we can set the reaction term to be the gradient of a data
term, i.e. $\psi(u) = \nabla_u \cD(u)$. For example, if $\cD(u, f_n) =
\frac{\lambda}{2} \|A u - f_n\|_2^2$, $\psi(u) = \lambda A^\top(Au -
f_n)$, where $f_n$ is the degraded image, $A$ is the
associated linear operator, and $\lambda$ is related to the strength
of the reaction term. In the case of Gaussian denoising, $A$ is the
identity matrix.

In our work, instead of making use of the well-chosen filters and influence functions, we train the nonlinear diffusion process 
for specific image restoration problem, including both the linear filters and the influence functions. As the diffusion process is 
an iterative approach, typically we run it for certain iterations. In order to make our proposed diffusion process more 
flexible, we train the parameters of the diffusion model for each single iteration. Finally, we arrive at a diffusion process which 
merely involves several iterations (referred to as stages). 

\subsection{Relations to existing image restoration models}
Previous works \cite{relations,biased} show that in the nonlinear
diffusion framework, there exist natural relations between reaction
diffusion and regularization based energy functional. First of all, we
can interpret~\eqref{ourreactionmodel} as one gradient descent step at
$u_{t-1}$ of a certain energy functional given by
\vspace*{-0.2cm}
\begin{equation}\label{foemodel}
E(u, f_n) = \suml{i = 1}{N_k} \mathcal{R}_i(u) + \cD(u, f_n)\,,
\vspace*{-0.2cm}
\end{equation}
where $\mathcal{R}_i(u) = \sum\nolimits_{p = 1}^{N} \rho_i^t((K_i^t
u)_p)$ are the regularizers and the functions $\rho_i^t$ are the so-called penalty
functions. Note that $\rho'(z) = \phi(z)$. Since the parameters \{$K_i^t,
\rho_i^t$\} vary across the stages, \eqref{foemodel} is a dynamic energy
functional, which changes at each iteration.

In the case of fixed \{$K_i^t, \rho_i^t$\} across the stages $t$, it
is obvious that functional \eqref{foemodel} is exactly the fields of
experts image prior regularized variational model for image
restoration \cite{RothFOE2009, ChenRP14, ChenPRB13}.  In our work, we
do not exactly solve this minimization problem anymore, but in
contrast, we run the gradient descent step for several stages, and
each gradient descent step is optimized by training.

In a very recent work \cite{CSF2014}, Schmidt \etal exploited an additive form of half-quadratic optimization to solve the 
same problem \eqref{foemodel}, which finally leads to a fast and effective image restoration model called 
cascade of shrinkage fields (CSF). 
The CSF model makes an assumption that the data term in \eqref{foemodel} is quadratic and the operator $A$ can 
be interpreted as a convolution operation, such that the corresponding subproblem has fast closed-form solution based on 
discrete Fourier transform (DFT). This restrains its applicability to many other problems such as image super resolution. However, our 
proposed diffusion model does not have this restriction on the data term. 
In principle, any smooth data term is appropriate. Moreover, as shown in the following sections, we can even handle the case of 
non-smooth data term. 

There exist some previous works \cite{Barbu2009, DomkeAISTATS2012}, also 
trying to train an optimized gradient descent algorithm for the energy functional similar to \eqref{foemodel}. 
In their works, the Gaussian denoising problem is considered, and 
the trained gradient descent algorithm typically involves less than 10 iterations. 
However, their model is much more constrained, in the sense that, they exploited the same 
filters for each gradient descent step. More importantly, the influence function in their model is fixed to be a unique one. 
This clearly restricts the model performance, as demonstrated in Sec. \ref{sec:denoising}. 

There are also few preliminary works, \eg, \cite{liu2010learning} 
to go beyond traditional PDEs of the form \eqref{pmmodel}, 
and propose to learn optimal PDEs for image restoration via optimal control. 
However, the investigated PDE model in \cite{liu2010learning} is too simple to 
generate a promising performance, as they only optimize the linear combination coefficients of a few predefined terms, 
which depend on selected derivative filters. 

\begin{figure}[t!]
\centering
\vspace*{-1.2cm}
\hspace*{-0.8cm} {\includegraphics[width=1.15\linewidth]{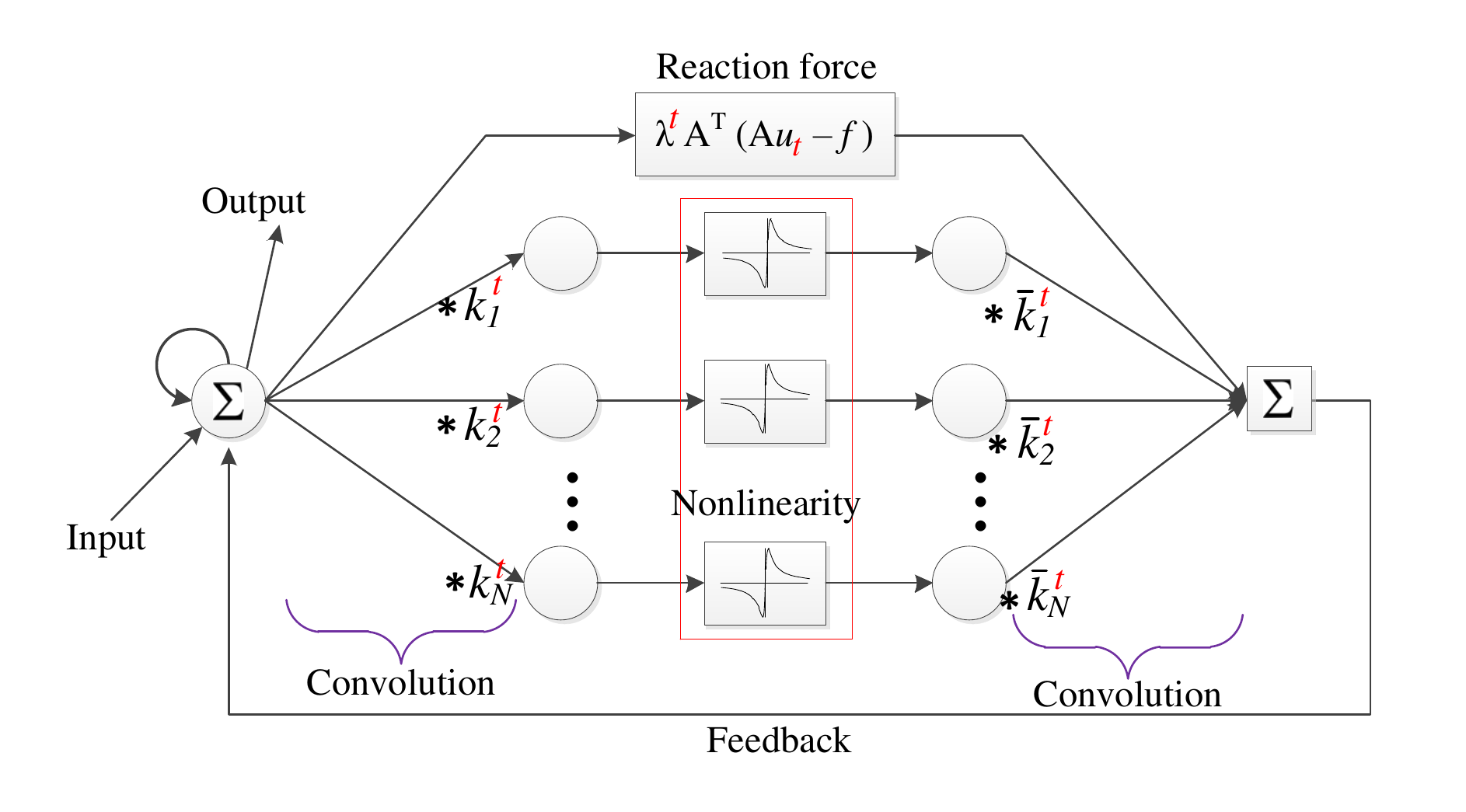}}
\vspace*{-0.75cm}
\caption{The architecture of our proposed diffusion model. Note that the additional convolution step with the rotated kernels $\bar k_i$
(\textit{cf.} Equ. \ref{denoising}) does not appear in conventional feed-forward CNs. Our model can be interpreted as a CN with a 
feedback step, which makes it different from conventional feed-forward networks. Due to the feedback step, 
it can be categorized into recurrent neural networks \cite{graves2009offline}. 
}\label{fig:recurrentCNN}
\vspace{-0.4cm}
\end{figure}

The proposed diffusion model also bears an interesting link to the convolutional networks (CNs) employed for 
image restoration problems \cite{CNNdenoising}. One can see that 
each iteration (stage) of our proposed diffusion process involves the convolution operation with a set of linear filters, 
and thus it can be treated as a convolutional network. The architecture of our proposed network is shown in Figure 
\ref{fig:recurrentCNN}, where one can see that it is not a pure feed-forward network any more, because it has a feedback step. 
Therefore, the structure of our CN model is different from conventional feed-forward networks. Due to this feedback step, 
it can be categorized into recurrent neural networks \cite{graves2009offline}. 
Moreover, the nonlinearity (\eg, influence functions in the context of nonlinear diffusion) 
in our proposed network are trainable. However, conventional CNs make use of 
fixed activation function, \eg, ReLU functions or sigmoid functions. 
\section{Learning}
In this paper, we train our models for two representative image restoration problems: (1) denoising of images 
corrupted by Gaussian noise and (2) JPEG blocking artifacts reduction, which is formulated as a non-smooth problem. 
We use a loss minimization scheme to learn the model parameters $\Theta_t = \{\lambda^t, \phi_i^t, k_i^t\}$ for each 
stage $t$ of the diffusion process, given $S$ training samples $\{f_n^{(s)},u_{gt}^{(s)}\}_{s=1}^S$, where 
$f_n^{(s)}$ is a noisy input and $u_{gt}^{(s)}$ is the corresponding ground truth clean image.

We firstly consider a greedy training strategy to train the diffusion processes stage-by-stage, \ie, at stage $t$, we minimize the 
cost function
\vspace*{-0.2cm}\begin{equation}\label{greedy}
\cL(\Theta_t ) = \suml{s = 1}{S}\ell(u_t^{(s)}, u_{gt}^{(s)}) \,,
\vspace*{-0.2cm}
\end{equation}
where $u_t^{(s)}$ is the output of stage $t$ of the diffusion process. 
We prefer the usual quadratic loss function to the negative PSNR used in \cite{CSF2014}, because 
the latter one imposes more weights on those samples with relatively smaller cost, and thus leads to slightly inferior results in practice. 
The loss function is given as
\begin{equation}\label{loss}
\ell(u_t^{(s)}, u_{gt}^{(s)}) = \frac 1 2 \|u_t^{(s)} - u_{gt}^{(s)}\|^2_2 \,.
\end{equation}
\textbf{Parameterizing the influence functions $\phi_i^t$}: We parameterize the influence function via standard radial basis 
functions (RBFs), \ie, each function $\phi$ is represented as a weighted linear combination of a family of RBFs as follows
\begin{equation}\label{rbf}
\phi_i^t(z) = \suml{j = 1}{M}w_{ij}^t\varphi \left(\frac {|z - \mu_j|}{\gamma_j}\right) \,,
\end{equation}
where $\varphi$ represents different RBFs. 
In this paper, we exploit RBFs with equidistant centers $\mu_j$ and unified scaling $\gamma_j$. We investigate two typical 
RBFs \cite{hu2010handbook}: (1) Gaussian radial basis and (2) triangular-shaped radial basis. 

In general, the Gaussian RBF can provide better approximation for generally smooth function than the 
triangular-shaped RBF with the same number of basis functions. However, the triangular-shaped RBF based function 
parameterization has the advantage of computational efficiency. More details can be found in the \textit{supplemental material} . 
In our work, we consider both function parameterization methods, but only present the results achieved based on 
the Gaussian RBF. 

\noindent \textbf{Training for denoising}: 
According to the diffusion equation \eqref{ourreactionmodel}, for image denoising, the output of stage $t$ is given as
\vspace*{-0.2cm}\begin{equation}\label{denoising}
u_t = u_{t-1} - \left(\sum\limits_{i = 1}^{N_k}\bar k_i^t * \phi_i^t(k_i^t * u_{t-1}) + \lambda^t (u_{t-1} - f_n)
\right)\,,
\vspace*{-0.2cm}
\end{equation}
where we explicitly use a convolution kernel $\bar k_i$ (obtained by rotating the kernel $k_i$ 180 degrees) 
to replace the $K_i^\top$ for the sake of model simplicity \footnote{We use the symmetric boundary condition in our work. In
this case, $K_i^\top$ can be interpreted as the convolution kernel $\bar k_i$ only in the central region. Therefore, we actually slightly 
modify the original model.}. 

\noindent \textbf{Training for deblocking:} Motivated by \cite{BrediesH12}, 
we consider a new variational model for JPEG deblocking based on the FoE image prior model 
\begin{equation}\label{vbdeblocking}
\arg\min\limits_{u}E(u) = 
\suml{i=1}{N_k}\rho_i(k_i * u) + \cI_{Q} (Du)\,,
\end{equation}
where $\cI_{Q}$ is a indicator function over the set $Q$ (quantization constraint set). In JPEG compression, information loss 
happens in the quantization step, where all possible values in the range $[d - 0.5, d + 0.5]$ ($d$ is an integer) are quantized to a 
single number $d$. Given a compressed data, we only know $d$. Therefore, all possible values in the interval $[d - 0.5, d + 0.5]$ 
define a convex set $Q$ which is a box constraint. The sparse matrix 
$D \in \R^{N \times N}$ denotes the block DCT transform. We refer to \cite{BrediesH12} for more details. 

We derive the diffusion process w.r.t the variational model \eqref{vbdeblocking} 
using the proximal gradient method \cite{ipiano}, which reads as
\begin{equation}\label{deblocking}
u_t = D^\top \text{proj}_{Q}\left(D \left(u_{t-1} - \sum\nolimits_{i = 1}^{N_k}\bar k_i^t * \phi_i^t(k_i^t * u_{t-1})\right) \right) \,,
\end{equation}
where $\text{proj}_{Q}(\cdot)$ denotes the orthogonal projection onto $Q$. More details can be found in the 
\textit{supplemental material} . 

\noindent \textbf{Gradients:} 
We minimize \eqref{greedy} with commonly used gradient based L-BFGS algorithm \cite{lbfgs}. The gradient of 
the loss function at stage $t$ w.r.t the model parameters $\Theta_t$ is computed using standard chain rule, given as
\begin{equation}\label{chainrule}
\frac {\partial \ell(u_t, u_{gt})}{\partial \Theta_t} = 
\frac {\partial u_t}{\partial \Theta_t} \cdot \frac {\partial \ell(u_t, u_{gt})}{\partial u_t} \,,
\end{equation}
where $\frac {\partial \ell(u_t, u_{gt})}{\partial u_t} = u_t - u_{gt}$ is directly derived from \eqref{loss}, 
$\frac {\partial u_t}{\partial \Theta_t}$ is computed from \eqref{denoising} for the training of denoising task or 
\eqref{deblocking} for the deblocking training, respectively. We do not present the derivatives for specific model parameters due 
to space limitation. All derivatives can be found in the \textit{supplemental material} . 

\noindent \textbf{Joint training:} In \eqref{greedy}, each stage is trained greedily such that the output of each stage is 
optimized according to the loss function, regardless of the total stages $T$ used in the diffusion process. A better strategy would be 
to jointly train all the stages simultaneously. The joint training task is formulated as 
\vspace*{-0.3cm}\begin{equation}\label{joint}
\cL(\Theta_{1, \cdots, T} ) = \suml{s = 1}{S}\ell(u_T^{(s)}, u_{gt}^{(s)}) \,,
\vspace*{-0.2cm}
\end{equation}
where the loss function only depends on $u_T$ (the output of the final stage $T$). The gradients of the loss function w.r.t 
$\Theta_t$ is given as
\[
\frac {\partial \ell(u_T, u_{gt})}{\partial \Theta_t} = 
\frac {\partial u_t}{\partial \Theta_t} \cdot \frac {\partial u_{t+1}}{\partial u_{t}} \cdots
\frac {\partial \ell(u_T, u_{gt})}{\partial u_T} \,,
\]
which is the standard back-propagation technique widely used in the neural networks learning \cite{lecun1998gradient}. Compared 
with the greedy training, we additionally need to calculate $\frac {\partial u_{t+1}}{\partial u_{t}}$. 
All the derivations can be found in the \textit{supplemental material} . 
\section{Experiments}
We used the same 400 training images as 
\cite{CSF2014}, and cropped a $180 \times 180$ region from each image, resulting in a total of 400 training samples 
of size $180 \times 180$, \ie, roughly 13 million pixels. 

We trained the proposed diffusion process with at most 8 stages to observe its saturation behavior after some stages. 
We first greedily trained $T$ stages of our model with 
specific model capacity, then conducted a joint training for the parameters of the whole $T$ stages. 

In our work, we mainly considered two trained reaction diffusion (TRD) models. 
\begin{subequations}
\begin{align*}
        \text{TRD}_{5 \times 5}^T, &\text{Fully trained model with 24 filters of size} ~5 \times 5\,,\\
        \text{TRD}_{7 \times 7}^T, &\text{Fully trained model with 48 filters of size} ~7 \times 7\,,
\end{align*}
\end{subequations}
where $\text{TRD}_{m \times m}^T$ denotes a nonlinear diffusion process of stage $T$ with filters of size $m \times m$. The 
filters number is $m^2 - 1$, if not specified. 

Note that the calculation of the gradients of the loss function in \eqref{chainrule} 
can be accomplished with convolution technique efficiently, even with a simple Matlab implementation. The training time 
varies greatly for different configurations. Important factors include (1) model capacity, (2) number of training samples, 
(3) number of iterations taken by the L-BFGS, and (4) number of Gaussian RBF kernels used for function approximation. 
We report below the most time consuming cases. 

In training, computing the gradients $\frac{\partial \cL}{\partial \Theta}$ with respect to the parameters of one stage 
for 400 images of size $180 \times 180$ takes about 35s ($\text{TRD}_{5 \times 5}$), 75s ($\text{TRD}_{7 \times 7}$) or 
165s ($\text{TRD}_{9 \times 9}$) with Matlab implementation on a server with CPUs: Intel(R) Xeon E5-2680 @ 2.80GHz 
(eight parallel threads, 63 Gaussian RBF kernels for the influence function parameterization). We typically run 200 L-BFGS 
iterations for optimization. Therefore, the total training time, \eg, for the $\text{TRD}^5_{7 \times 7}$ model is about 
$5 \times (200 \times 75)/3600 = 20.8h$. \textit{Code for learning and inference is available on the authors' homepage 
\url{www.GPU4Vision.org}.  }

\subsection{Image denoising experiments}\label{sec:denoising}
We started with the training model of $\text{TRD}_{5 \times 5}^T$. We first considered the greedy scheme to train a 
diffusion process up to 8 stages (\ie, $T \leq 8$), in order to observe the asymptotic behavior of the diffusion process. 
After the greedy training was completed, we conducted joint training for a diffusion model of certain stages 
(\eg, $T = 5$), by simultaneously tuning the parameters in all stages. 

We initialized the joint training with the parameters 
obtained from greedy training, as this is guaranteed not to decrease the training performance. 
In previous work \cite{CSF2014}, it is shown that joint training 
a model with filters of size $5 \times 5$ or larger hardly makes a difference relative to the result obtained by the greedy training. 
However, in our work we observed that joint training always improves the result of greedy training. 

Note that for the models trained in the greedy manner, 
we can stop the inference at any stage, as its output of each stage is optimized. However, for the jointly trained models, we have to 
run $T$ stages, as in this case only the output of the $T^{th}$ stage is optimized. 

We first trained our diffusion models for the Gaussian denoising problem with standard deviation $\sigma = 25$. The 
noisy training images were generated by adding synthetic Gaussian noise with $\sigma = 25$ to the clean images. 
Once we have trained a diffusion model, we evaluated its performance on a standard test dataset of 
68 natural images.\footnote{The test images are strictly separate from the training datasets.}

We present the final results of the joint training 
in Table \ref{denoisingresults}, together with a selection of recent state-of-the-art denoising algorithms, 
namely BM3D \cite{BM3D}, LSSC \cite{LSSC}, EPLL \cite{EPLL}, opt-MRF \cite{ChenPRB13}, $\text{RTF}_5$ model 
\cite{ECCV2012RTF} and two very recent methods: the CSF model \cite{CSF2014} and WNNM \cite{WNNM}. We downloaded 
these algorithms from the corresponding author's homepage, and used them as is.

\begin{table}[t!]
\centering
\begin{tabular}{l c c c l l}
\Xhline{0.5pt}
\cline{1-6}
\multirow{2}*{Method} & \multicolumn{2}{c}{$\sigma$} &\multirow{2}*{\color{blue}St.} 
& \multicolumn{2}{c}{${\sigma = 15}$}\\
\cline{2-3}
\cline{5-6}
&15 & 25& &$\text{TRD}_{5 \times 5}$ & $\text{TRD}_{7\times 7}$\\
\cline{1-6}
BM3D &31.08 &28.56 &\cellcolor[gray]{0.70}\color{blue}2 & \cellcolor[gray]{0.90}31.14 & \cellcolor[gray]{0.90}31.30\\
LSSC&31.27 &28.70 &\cellcolor[gray]{0.70}\color{blue}5 & \cellcolor[gray]{0.90}31.30 & \cellcolor[gray]{0.90}\textbf{31.42}\\
EPLL&31.19 &28.68 &\cellcolor[gray]{0.70}\color{blue}8 & \cellcolor[gray]{0.90}31.34 & \cellcolor[gray]{0.90}\textbf{31.43}\\
\cline{4-6}
opt-MRF&31.18 &28.66 & & \multicolumn{2}{c}{$\sigma = 25$}\\
\cline{5-6}
$\text{RTF}_5$ &-- &28.75 & & $\text{TRD}_{5 \times 5}$ & $\text{TRD}_{7\times 7}$\\
WNNM&{31.37} &{28.83} &\cellcolor[gray]{0.70}\color{blue}2 & \cellcolor[gray]{0.90}28.58& \cellcolor[gray]{0.90}28.77\\
$\text{CSF}_{5 \times 5}^5$&31.14 &28.60 &\cellcolor[gray]{0.70}\color{blue}5 & \cellcolor[gray]{0.90}28.78 & 
\cellcolor[gray]{0.90}\textbf{28.92}\\
$\text{CSF}_{7 \times 7}^5$&31.24 &28.72 &\cellcolor[gray]{0.70}\color{blue}8 
& \cellcolor[gray]{0.90}{28.83} & \cellcolor[gray]{0.90}\textbf{28.95}\\
\Xhline{0.5pt}
\cline{1-6}
\end{tabular}
\vspace*{0.2cm}
\caption{Average PSNR (dB) on 68 images from \cite{RothFOE2009} for image denoising with $\sigma = 15, 25$.}
\label{denoisingresults}
\vspace*{-0.5cm}\end{table}

Concerning the performance of our $\text{TRD}_{5 \times 5}^T$ models, we find that joint training usually leads to 
an improvement of about 0.1dB in the cases of $T \geq 5$. From Table \ref{denoisingresults}, one can see that 
(1) the performance of the $\text{TRD}_{5 \times 5}^T$ model saturates 
after stage 5, \ie, in practice, 5 stages are typically enough; (2) our 
$\text{TRD}_{5 \times 5}^5$ model has achieved significant improvement (28.78 \vs 28.60), 
compared to a similar model $\text{CSF}_{5 \times 5}^5$, which has the same model capacity and 
(3) moreover, our $\text{TRD}_{5 \times 5}^8$ model is on par with so far the best-reported algorithm - WNNM. 

When comparing with some closely related models 
such as the FoE prior based variational model \cite{ChenPRB13}, the FoE derived CSF model 
\cite{CSF2014} and convolutional networks (CNs) \cite{CNNdenoising}, 
our trained models can provide significantly superior performance. Therefore, 
a natural question arises: what is the critical factor in the effectiveness 
of the trained diffusion models? There are actually two main aspects in 
our training model: (1) the linear filters and (2) the influence functions. In order to have a better understanding of the trained 
models, we went through a series of experiments to investigate the impact of these two aspects. 

\noindent \textbf{Analysis of the proposed diffusion process:} 
Concentrating on the model capacity of 24 filters of size $5 \times 5$, 
we considered the training of a diffusion process with 10 steps, \ie, $T = 10$ for the Gaussian denoising of noise level $\sigma = 25$. 
We exploited two main classes of configurations: 
({\color{blue}{A}}) the parameters of every stage are the same and 
({\color{blue}{B}}) every diffusion stage is different from each other. In both configurations, 
we consider two cases: ({\color{blue}{I}}) only train the linear filters with fixed influence 
function $\phi(z) = 2z/(1+z^2)$ and ({\color{blue}{II}}) simultaneously 
train the filters and influence functions. 

Based on the same training dataset and test dataset, we obtained the following results: 
({\color{blue}{A}}.{\color{blue}{I}}) every diffusion step is the same, and only the filters are optimized with fixed influence 
function. This is a similar configuration to previous works \cite{Barbu2009, DomkeAISTATS2012}. 
The trained model achieves a test performance of 28.47dB. 
({\color{blue}{A}}.{\color{blue}{II}}) with additional tuning of the influence functions, the resulting performance  is boosted to 
28.60dB. ({\color{blue}{B}}.{\color{blue}{I}}) every diffusion step can be different, but only train the linear filters with fixed 
influence function. The corresponding model obtains the result of 28.56dB, which is equivalent to 
the variational model \cite{ChenPRB13} with the same model capacity. Finally ({\color{blue}{B}}.{\color{blue}{II}}) 
with additional optimization of the influence functions, the trained model leads to a significant improvement with the result of 
28.86dB. 

The analysis experiments demonstrate that without the training of the influence functions, there is no chance to achieve 
significant improvements over previous works, 
no matter how hard we tune the linear filters. Therefore, we believe that the additional freedom to tune the 
influence functions is the critical factor of our proposed training model. 
After having a closer look at the learned influence functions of the $\text{TRD}_{5 \times 5}^5$ model, these functions 
reinforce our argument. 

\noindent \textbf{Learned influence functions:} The form of 120
learned penalty functions $\rho$\footnote{The penalty function $\rho(z)$ is integrated from the influence function $\phi(z)$ according 
to the relation $\phi(z) = \rho'(z)$} in the $\text{TRD}_{5 \times 5}^5$
model can be divided into four classes (see the corresponding
subfigures in Figure~\ref{functions}):
\begin{itemize}
\setlength\itemsep{-0.5em}
\item[(a)] Truncated convex penalty functions with low values around
  zero to encourage smoothness.
\item[(b)] Negative Mexican hat functions, which have a local minimum
  at zero and two symmetric local maxima.
\item[(c)] Truncated concave functions with smaller values at the two
  tails.
\item[(d)] Double-well functions, which have a local maximum (not a
  minimum any more) at zero and two symmetric local minima.
\end{itemize}
At first glance, the learned penalty functions (except (a)) differ
significantly from the usually adopted penalty functions used in PDE
and energy minimization methods. However, it turns out that they have
a clear meaning for image regularization.

Regarding the penalty function (b), there are two critical points (indicated by red triangles). When the magnitude of the 
filter response is relatively small (\ie, less than the critical points), 
probably it is stimulated by the noise and therefore the penalty function encourages 
smoothing operation as it has a local minimum at zero. However, once the magnitude of the filter response is large enough 
(\ie, across the critical points), the corresponding local patch probably contains a real image edge or certain structure. 
In this case, the penalty function encourages to increase the magnitude of the filter response, alluding to an image sharpening 
operation. Therefore, the diffusion process controlled by the influence function (b), can adaptively switch between 
image smoothing (forward diffusion) and sharpening (backward diffusion). We find that the learned influence function (b) is closely 
similar to an elaborately designed function in a previous work \cite{gilboa2002forward}, 
which leads to an adaptive forward-and-backward diffusion process. 

A similar penalty function to the learned function (c) with a concave shape 
is also observed in previous work on image prior learning \cite{zhu1997prior}. This penalty function also encourages to 
sharpen the image edges. Concerning the learned penalty function (d), as it has local minima at two specific points, it 
prefers specific image structure, implying that it helps to form certain image structure. We also find that this penalty function is 
exactly the type of bimodal expert functions for texture synthesis employed in \cite{HeessWH09}. 

Now it is clear that the diffusion process involving the learned influence functions does not perform 
pure image smoothing any more for image processing. 
In contrast, it leads to a diffusion process for adaptive image smoothing 
and sharpening, distinguishing itself from previous commonly used image regularization techniques. 
\begin{figure}[t!]
\centering
\subfigure[48 filters of size $7 \times 7$ in stage 1]{\includegraphics[width=1\linewidth]{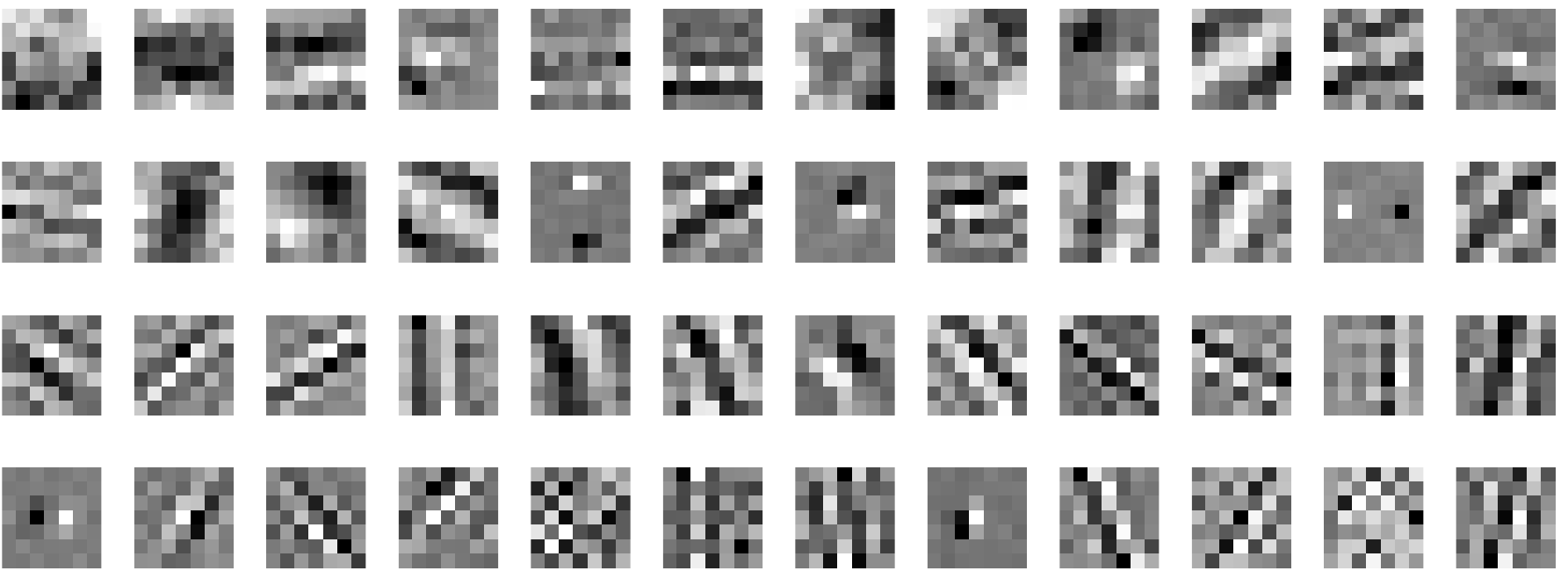}}\\
\vspace*{-0.25cm}
\subfigure[48 filters of size $7 \times 7$ in stage 5]{\includegraphics[width=1\linewidth]{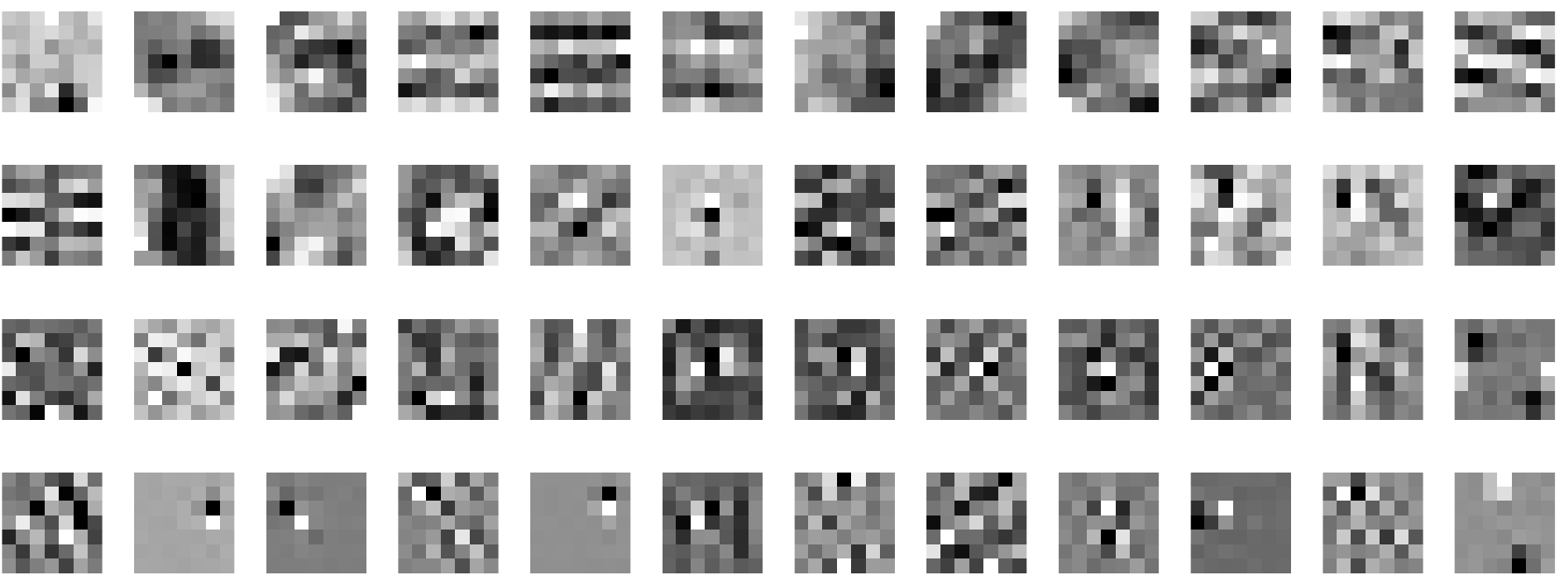}}
\caption{Trained filters (in the first and last stage) of the $\text{TRD}^5_{7\times 7}$ model for the noise level $\sigma = 25$. 
We can find first, second and higher-order derivative filters, as well as rotated derivative filters along different directions. These filters 
are effective for image structure detection, such as image edge and texture.}
\label{fig:filters}
\vspace*{-0.5cm}
\end{figure}

\noindent \textbf{Influence of initialization:} 
Our training model is also 
a deep model with many stages (layers). It is well-known that deep models are usually sensitive to initialization. 
However, our training model is not very sensitive to initialization. 
We have training experiments with fully random initializations for 
(1) greedy training. Using fully random initial parameters in range $[-0.5, 0.5]$, the trained models lead to a 
deviation within 0.01dB in the test phase and (2) joint training. Fully random initializations lead to models with inferior 
results, e.g., $\text{TRD}^5_{5\times 5}$ (28.61 \vs 28.78). 
However, a plain initialization (all stages with DCT filters, influence function $\phi(z) = 2z/(1+z^2)$) 
works almost the same, e.g.,$\text{TRD}^5_{5\times 5}$, (28.75 \vs 28.78) and 
$\text{TRD}^5_{7\times 7}$, (28.91 \vs 28.92). 

\begin{table}[t!]
\centering
\hspace*{-0.2cm} \begin{tabular}{c c c c c c}
\Xhline{0.5pt}
\cline{1-6}
Method & $256^2$ & $512^2$ & $1024^2$ & $2048^2$ & $3072^2$\\
\cline{1-6}
\rowcolor[gray]{0.85} BM3D \cite{BM3D} &1.1 &4.0 &17 & 76.4 & {\color{black}{176.0}}\\
\rowcolor[gray]{0.85} $\text{CSF}_{7 \times 7}^5$ \cite{CSF2014} &3.27 &11.6 &40.82 & 151.2 & {\color{black}{494.8}}\\
\rowcolor[gray]{0.85} WNNM \cite{WNNM} &122.9 & 532.9 &2094.6 & -- & --\\
\cline{1-6}
\multirow{3}*{$\text{TRD}^5_{5 \times 5}$} & \cellcolor[gray]{0.85} 0.51 & 
\cellcolor[gray]{0.85} 1.53 & \cellcolor[gray]{0.85} 5.48 
&\cellcolor[gray]{0.85} 24.97 & \cellcolor[gray]{0.85} 53.3\\
& {\color{blue}{0.43}}& {\color{blue}{0.78}}& {\color{blue}{2.25}}& {\color{blue}{8.01}}& {\color{blue}{21.6}}\\
& {\color{red}{0.005}} & {\color{red}{0.015}} & {\color{red}{0.054}} &{\color{red}{0.18}} & {\color{red}{0.39}}\\
\cline{1-6}
\multirow{3}*{$\text{TRD}^5_{7 \times 7}$} & \cellcolor[gray]{0.85} 1.21 & 
\cellcolor[gray]{0.85} 3.72 & \cellcolor[gray]{0.85} 14.0 
&\cellcolor[gray]{0.85} 62.2& \cellcolor[gray]{0.85} 135.9\\
& {\color{blue}{0.56}}& {\color{blue}{1.17}}& {\color{blue}{3.64}}& {\color{blue}{13.01}}& {\color{blue}{30.1}}\\
& {\color{red}{0.01}} & {\color{red}{0.032}} &{\color{red}{0.116}} & {\color{red}{0.40}} & {\color{red}{0.87}}\\
\Xhline{0.5pt}
\cline{1-6}
\end{tabular}
\vspace*{0.2cm}
\caption{Run time comparison for image denoising (in seconds) with different implementations. 
(1) The run time results with \colorbox{grayB}{gray}background are evaluated 
with the single-threaded implementation on Intel(R) Xeon(R) CPU E5-2680 v2 @ 2.80GHz; 
(2) the {\color{blue}{blue}} colored run times are obtained with multi-threaded computation using Matlab \textit{parfor} 
on the above CPUs; (3) the run time results colored in {\color{red}{red}} are executed on a NVIDIA GeForce GTX 780Ti GPU. 
We do not count the memory transfer time between CPU/GPU for the GPU implementation (if counted, the run time will nearly 
double)}\label{run time}
\vspace*{-0.5cm}\end{table}
\noindent \textbf{Training for other configurations:}
In order to investigate the influence of the model capacity, we increase the filter size to $7 \times 7$ and $9 \times 9$. We find 
that increasing the filter size from $5 \times 5$ to $7 \times 7$ brings a significant improvement of 0.14dB (
$\text{TRD}^5_{7\times 7}$ \vs $\text{TRD}^5_{5\times 5}$) as show in Table \ref{denoisingresults}. 
However, if we further increase the filter size to $9 \times 9$, the resulting $\text{TRD}^5_{9\times 9}$ leads to a 
performance of 28.96dB (a slight improvement of 0.05dB relative to the $\text{TRD}^5_{7\times 7}$ model). In 
practice, we prefer the $\text{TRD}^5_{7\times 7}$ model as it provides the best trade-off between performance and 
computation time. Fig. \ref{fig:filters} shows the trained filters of the $\text{TRD}^5_{7\times 7}$ model in the 
first and last stage. 

We also trained diffusion models for the noise level of $\sigma = 15$, and the test performance is shown in 
Table \ref{denoisingresults}. In experiments, we observed that joint training can always gain an improvement of about 0.1dB 
over the greedy training for the cases of $T \geq 5$. 

From Table \ref{denoisingresults}, one can see that for both noise levels, 
the resulting $\text{TRD}_{7\times 7}$ model achieves the highest average PSNR. 
The $\text{TRD}^5_{7\times 7}$ model outperforms the benchmark 
BM3D method by 0.35dB in average. This is a notable improvement as few methods can surpass BM3D more 
than 0.3dB in average \cite{levin2011natural}. Moreover, the $\text{TRD}^5_{7\times 7}$ model also surpasses 
the best-reported algorithm - WNNM method, which is very slow as shown in Table \ref{run time}. 
In summary, our $\text{TRD}^5_{7\times 7}$ model outperforms all the recent state-of-the-arts on the 
exploited test dataset, meanwhile it is the fastest method even with the CPU implementation. 

\noindent \textbf{Run time:} The algorithm structure of our TRD model is closely similar to the CSF model, which is well-suited 
for parallel computation on GPUs. We implemented our trained models on GPU using CUDA programming to 
speed up the inference procedure, and finally it indeed lead to a significantly improved run time, see Table \ref{run time}. 
We see that for the images of size up to $3\text{\normalfont K} \times 3\text{\normalfont K}$, 
the $\text{TRD}^5_{7\times 7}$ model is still able to  
accomplish the denoising task in less than 1s. 

We make a run time comparison to other denoising algorithms based on strictly enforced single-threaded CPU computation (
\eg, start Matlab with -singleCompThread) for a fair comparison, see Table \ref{run time}. We only 
present the results of some selective algorithms, which either have the best denoising result or run time performance. 
We refer to \cite{CSF2014} for a comprehensive run time comparison of various algorithms\footnote{
LSSC, EPLL, opt-MRF and $\text{RTF}_5$ methods are much slower than BM3D on the CPU, \textit{cf.} \cite{CSF2014}. }.

We see that our TRD model is generally faster than the CSF model with the same model capacity. 
It is reasonable, because in each stage the CSF model involves additional DFT and inverse DTF operations, \ie, 
our model only requires a portion of the computation of the CSF model. Even though the BM3D is a non-local model, 
it still possesses high computation efficiency. In contrast, another non-local model - WNNM achieves compelling 
denoising results but at the expense of huge computation time. Moreover, the WNNM algorithm is hardly applicable for 
high resolution images (\eg, 10 mega-pixels) due to its huge memory requirements. 
Note that our model can be also easily implemented with multi-threaded CPU computation. 


\subsection{JPEG deblocking experiments}
\begin{table}[t!]
\centering
\hspace*{-0.15cm} \begin{tabular}{|l|L{0.8cm}|C{0.8cm}|C{0.9cm}|C{0.8cm}|C{0.9cm}|C{0.9cm}|}
\hline
$q$ & {\small JPEG \newline decoder} & \small TGV \newline \cite{BrediesH12}& \small Dic-SR\cite{TVdeblockingDic}
&\footnotesize SADCT \cite{foi2007pointwise} &\small RTF\cite{ECCV2012RTF} & \small $\text{TRD}^4_{7 \times 7}$\\
\hline\hline
10 & 26.59 & 26.96 & 27.15 & 27.43 & {27.68} & \textbf{27.85}\\
\hline
20 & 28.77 & 29.01 & 29.03 & 29.46 & {29.83} & \textbf{30.06}\\
\hline
30 &  30.05 & 30.25 & 30.13 & 30.67 & {31.14} & \textbf{31.41}\\
\hline
\end{tabular}
\vspace*{0.2cm}
\caption{JPEG deblocking results for natural images, reported with average PSNR values.}
\label{results}
\vspace*{-0.5cm}\end{table}
We also trained diffusion models for the JPEG deblocking problem. 
We followed the test procedure in \cite{ECCV2012RTF} for performance evaluation. 
The test images were converted to gray-value, and scaled by a factor 
of 0.5, resulting images of size $240 \times 160$. We distorted the images by JPEG blocking artifacts. We considered 
three compression quality settings $q = 10, 20 ~\text{and} ~30$ for the JPEG encoder. 

We trained three nonlinear diffusion $\text{TRD}_{7 \times 7}$ models for different compression parameter $q$. 
We found that for JPEG deblocking, 4 stages are already enough. Results of the trained models are shown in Table 
\ref{results}, compared with several representative deblocking approaches. 
We see that our trained $\text{TRD}^4_{7 \times 7}$ outperforms all the competing approaches in terms of 
PSNR. Furthermore, our model is extremely fast on GPU, \eg, for a common image size of $1024 \times 1024$, our 
model takes about 0.095s, while the strongest competitor (in terms of run time) - SADCT 
consumes about 56.5s with CPU computation\footnote{
RTF is slower than SADCT, as it depends on the output of SADCT.}. 
See the \textit{supplemental material}  for JPEG deblocking examples.
\section{Conclusion and future work}
We have proposed a trainable reaction diffusion model for effective image restoration. Its critical point lies in 
the training of the influence functions. We have trained our models for 
the problem of Gaussian denoising and JPEG deblocking. Based on standard test datasets, 
the trained models result in the best-reported results. 
We believe that the effectiveness of the trained diffusion models is attributed to the following desired properties of the models
\begin{itemize}
\setlength\itemsep{0em}
    \item \textit{Anisotropy}. In the trained filters, we can find rotated derivative filters in different directions, \textit{cf.} Fig 
\ref{fig:filters}. 
    \item \textit{Higher order}. The learned filters contain first, second and higher-order derivative filters, \textit{cf.} Fig 
\ref{fig:filters}.  
    \item \textit{Adaptive forward/backward diffusion through the learned nonlinear functions}. Nonlinear functions corresponding to 
explicit backward diffusion appear in the learned nonlinearity, \textit{cf.} Fig \ref{functions}.
\end{itemize}

Meanwhile, the trained models are very simple and well-suited for parallel computation on GPUs. As a consequence, 
the resulting algorithms are significantly faster than all competing algorithms and hence are also applicable to the restoration 
of high resolution images. 

\noindent \textbf{Future work:} From a application point of view, we
think that it will be interesting to consider learned, nonlinear
reaction diffusion based models also for other image processing tasks
such as image super resolution, blind image deconvolution, optical
flow. Moreover, since learning the influence functions turned out to
be crucial, we believe that learning optimal nonlinearities in CNs
could lead to a similar performance increase. Finally, it will also be
interesting to investigate the unconventional penalty functions
learned by our approach in usual energy minimization approaches.

\section{Acknowledgments}
This work was supported by the Austrian Science Fund (FWF) under the China Scholarship Council 
(CSC) Scholarship Program and the START project BIVISION, No. Y729.

{\footnotesize
\bibliographystyle{ieee}
\bibliography{cvpr_bib}
}



\section*{Supplementary material}

# Notes on diffusion networks
–Supplemental Material for "On learning optimized reaction diffusion processes for effective image restoration"–

Yunjin Chen[1,2]
[1]Graz University of Technology   [2]National University of Defense Technology
chenyunjin_nudt@hotmail.com

## Abstract

*This is the supplemental material for our CVPR2015 paper entitled "On learning optimized reaction diffusion processes for effective image restoration". In this supplemental material, we give detailed derivations of gradients required in training for corresponding diffusion networks. In addition, we also present more image denoising and JPEG deblocking examples.*

## 1. Preliminaries

When we modify the original diffusion equation

$$u_t = u_{t-1} - \left(\sum_{i=1}^{N_k} K_i^{t\top} \phi_i^t(K_i^t u_{t-1}) + \lambda^t(u_{t-1} - f_n)\right), \tag{1}$$

to the following version

$$u_t = u_{t-1} - \left(\sum_{i=1}^{N_k} \bar{k}_i^t * \phi_i^t(k_i^t * u_{t-1}) + \lambda^t(u_{t-1} - f_n)\right), \tag{2}$$

we find it introduces some imperfections at the image boundary. The basic reason lies in the fact that, in the case of symmetric boundary condition used in our work, $K^\top v$ can be interpreted as the convolution with the kernel $\bar{k}$[1] only in the central region of image $v$. This interpretation does not hold for the image boundary. However, in the diffusion equation (2), the convolution kernel $\bar{k}_i$ is applied to the whole image, thus bringing some artifacts at the boundary. The benefit to use the diffusion equation (2), rather than (1) is that the revised model is more tractable in practice, especially for training, as everything can be done by the convolution operation efficiently.

In order to remove this artifacts, we pad the input image $u_{t-1}$ for stage $t$, as well as the noisy image $f_n$, with mirror reflections of itself. This operation is formulated by the sparse "padding" matrix $P$. After a diffusion step, we only crop the central region of the output image $u_t$ for usage. This operation is formulated by the sparse "cropping" matrix $T$. When we apply the matrix $P_T = P \times T$ to an image $u$, $P_T u$ corresponds to two operations: it first crops the central region of $u$, then pads it with mirror reflections.

After taking into account the operation of boundary handling, the exact diffusion process is illustrate in Figure 1. There we have $u_{tp} = P_T u_t$.

**In our derivations, we use the symmetric boundary condition for the convolution operation $k * u$ (image $u \in \mathbb{R}^{m \times n}$, $k \in \mathbb{R}^{r \times r}$). As we know, it is equivalent to the matrix-vector product formulation $Ku$, where $K \in \mathbb{R}^{N \times N}$ is a highly sparse matrix and $u$ is a column vector $u \in \mathbb{R}^N$ with $N = m \times n$. The result $k * u$ can also be interpreted with $Uk$, where matrix $U \in \mathbb{R}^{N \times R}$ is constructed from image $u$ and $k$ is a column vector $k \in \mathbb{R}^R$ with $R = r \times r$. This formulation is very helpful for the computation of the gradients of the loss function w.r.t. the kernel $k$, as $U^\top v$ ($v \in \mathbb{R}^N$ is a column vector) can be explicitly interpreted as a convolution operation, which is widely used in classic convolutional neural networks [1]. In the following derivations, we will make use of this equivalence frequently,** *i.e.*,

$$k * u \iff Ku \iff Uk\,.$$

---
[1] Recall that kernel $\bar{k}$ is obtained by rotating $k$ 180 degrees.

1

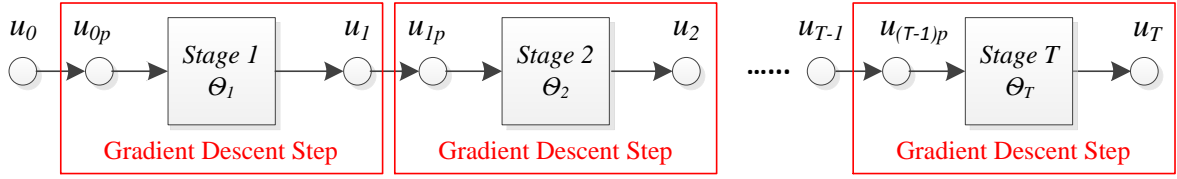

Figure 1. Proposed nonlinear diffusion process with careful boundary handling operation. Note that $u_{tp} = P_T u_t$.

The derivations also require matrix calculus. We use the **denominator layout notation** [2] for all the derivations.

## 2. Derivations of learning problem

Given $S$ training samples $\{f_n^{(s)}, u_{gt}^{(s)}\}_{s=1}^S$, where $f_n^{(s)}$ is the noisy input and $u_{gt}^{(s)}$ is the corresponding ground truth clean image. Let assume the original image size is $m \times n$. We first pad the noisy image $f_n$ with $\omega$ pixels, then the resulting image has size $O = (m + 2\omega) \times (n + 2\omega)$. We have the corresponding matrix $T \in \mathbb{R}^{N \times O}$, $P \in \mathbb{R}^{O \times N}$, $P_T \in \mathbb{R}^{O \times O}$ and $f_{np} \in \mathbb{R}^O, u_{gt} \in \mathbb{R}^N$.

### 2.1. Greedy training

In the greedy training for stage $t$, we are to minimize the following loss function w.r.t the model parameters $\Theta_t = \{\lambda^t, \phi_i^t, k_i^t\}$ of stage $t$,

$$L(\Theta_t) = \sum_{s=1}^S \ell(u_t^{(s)}, u_{gt}^{(s)}) = \sum_{s=1}^S \frac{1}{2}\|Tu_t^{(s)} - u_{gt}^{(s)}\|_2^2, \quad (3)$$

where

$$u_t^s = u_{(t-1)p}^s - \left(\sum_{i=1}^{N_k} \bar{k}_i^t * \phi_i^t(k_i^t * u_{(t-1)p}^s) + \lambda^t(u_{(t-1)p}^s - f_{np}^s)\right). \quad (4)$$

Note that in the training for stage $t$, the images $u_{(t-1)p}$ are fixed, served as the input of this feed-forward step.

As the gradient of overall loss function on the whole training datasets can be decomposed to the sum over training samples, in the following derivation, we only consider the case of one training sample for the sake of brevity. The basic result of the gradient of the loss function w.r.t the training parameters $\Theta_t = \{\lambda^t, \phi_i^t, k_i^t\}$ is given

$$\frac{\partial \ell(u_t, u_{gt})}{\partial \Theta_t} = \frac{\partial u_t}{\partial \Theta_t} \cdot \frac{\partial \ell(u_t, u_{gt})}{\partial u_t}, \quad (5)$$

where $\frac{\partial \ell(u_t, u_{gt})}{\partial u_t}$ is simply given as

$$\frac{\partial \ell(u_t, u_{gt})}{\partial u_t} = T^\top (Tu_t - u_{gt}).$$

Let us define a column vector $e \in \mathbb{R}^O$ as

$$e = T^\top (Tu_t - u_{gt}).$$

Therefore, the main issue is to calculate $\frac{\partial u_t}{\partial \Theta_t}$ from (4).

**Weight parameter $\lambda^t$:** It is easy to see that

$$\frac{\partial u_t}{\partial \lambda^t} = (u_{(t-1)p} - f_{np})^\top. \quad (6)$$

Thus $\frac{\partial \ell}{\partial \lambda^t}$ is given as

$$\boxed{\frac{\partial \ell}{\partial \lambda^t} = (u_{(t-1)p} - f_{np})^\top e.} \quad (7)$$

---

[2] http://en.wikipedia.org/wiki/Matrix_calculus

**Filters $k_i^t$:** Concerning the dependency of $u_t$ on parameters $k_i^t$, it is easy to see the following relationship

$$u_t^s \to \underbrace{-\bar{k}_i^t}_{f} * \underbrace{\phi_i^t(k_i^t * u_{(t-1)p})}_{v},$$

where $f$ and $v$ are two auxiliary variables defined as $f = -\bar{k}_i^t$ and $v = \phi_i^t(k_i^t * u_{(t-1)p})$. Therefore, we get the following dependency relationship,

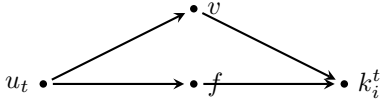

According to the chain rule, we have

$$\frac{\partial u_t}{\partial k_i^t} = \frac{\partial f}{\partial k_i^t} \cdot \frac{\partial u_t}{\partial f} + \frac{\partial v}{\partial k_i^t} \cdot \frac{\partial u_t}{\partial v}. \tag{8}$$

Note that $f = -\bar{k}_i^t$, which can be formulated as $f = -P_{inv}k_i^t$ with matrix $P_{inv}$ inverting the kernel vector $k_i^t$. Recall the equivalence

$$f * v \iff Fv \iff Vf.$$

Therefore, the first term of (8) is given as

$$\frac{\partial f}{\partial k_i^t} \cdot \frac{\partial u_t}{\partial f} = -P_{inv}^\top V^\top.$$

For the second term, we introduce an additional auxiliary variable $z$, defined as $z = k_i^t * u_{(t-1)p}$. Then we have $v = \phi_i^t(z)$. Recall that

$$z = k_i^t * u_{(t-1)p} \iff U_{(t-1)p} k_i^t.$$

Therefore, we obtain

$$\frac{\partial v}{\partial k_i^t} \cdot \frac{\partial u_t}{\partial v} = \frac{\partial z}{\partial k_i^t} \cdot \frac{\partial v}{\partial z} \cdot \frac{\partial u_t}{\partial v} = U_{(t-1)p}^\top \Lambda F^\top,$$

where $\Lambda$ is a diagonal matrix $\Lambda = \text{diag}(\phi_i^{t'}(z_1), \cdots, \phi_i^{t'}(z_p))$ ($\phi_i^{t'}$ is the first order derivative of function $\phi_i^t$). Note that $F = -\bar{K}_i^t$. In summary, $\frac{\partial u_t}{\partial k_i^t}$ is given as

$$\frac{\partial u_t}{\partial k_i^t} = -\left(P_{inv}^\top V^\top + U_{(t-1)p}^\top \Lambda \bar{K}_i^{t\top}\right). \tag{9}$$

Finally, we arrive at the desired gradients

$$\boxed{\frac{\partial \ell}{\partial k_i^t} = -\left(P_{inv}^\top V^\top + U_{(t-1)p}^\top \Lambda \bar{K}_i^{t\top}\right) e.} \tag{10}$$

In practice, we do not need to explicitly construct the matrices $V, U, \bar{K}_i^t$. Recall that the product of matrices $V^\top, U_{(t-1)p}^\top$ and a vector can be computed by the convolution operator [1]. As shown in a previous work [4], $\bar{K}_i^{t\top}$ can also be computed by the convolution operation with the kernel $\bar{k}_i^t$ with careful boundary handling. Matrix $P_{inv}^\top$ is merely a linear operation which inverts the vectorized kernel $k$. In the case of a square kernel $k$, it is equivalent to the Matlab command

$$P_{inv}^\top k \iff rot90(rot90(k)).$$

If we have a closer look at the diffusion equation (4), we find that it has a scaling problem w.r.t the filters $k_i^t$. First we know the function $\phi_i^t$ is free to tune in the training. In this case, if we scale the filter $k_i^t$ by a factor of $h$ to generate a new filter $\hat{k}_i^t = hk_i^t$, and the corresponding new function $\hat{\phi}_i^t$ is selected such that $\hat{\phi}_i^t(hz) = \frac{1}{h}\phi_i^t(z)$, then we will see that the term $\bar{k}_i^t * \phi_i^t(k_i^t * u_{(t-1)p})$ keep unchanged, *i.e.*, two different set of parameters $\{k_i^t, \phi_i^t\}$ and $\{\hat{k}_i^t, \hat{\phi}_i^t\}$ own exactly the same loss function $\ell(u_t, u_{gt})$. In order to get rid of this ambiguity, it is necessary to fix the scale of the filters. In practice, we learn filters with fixed unit norm. Motivated by the finding in [4] that meaningful filters should be zero-mean, we also construct the

training filter $k$ from the DCT basis $\mathcal{B} \in \mathbb{R}^{R \times (R-1)}$ (without the DC-component). Therefore, we define each filter $k \in \mathbb{R}^R$ with

$$k = \mathcal{B} \frac{c}{\|c\|_2}, \tag{11}$$

where $c \in \mathbb{R}^{R-1}$. Now the training parameters become $c$, and we need to calculate $\frac{\partial \ell}{\partial c}$. As shown in (10), we already have $\frac{\partial \ell}{\partial k}$, according to the chain rule, we have

$$\frac{\partial \ell}{\partial c} = \frac{\partial k}{\partial c} \cdot \frac{\partial \ell}{\partial k},$$

where $\frac{\partial k}{\partial c}$ is computed from (11). Let us define an auxiliary variable $v = \frac{c}{\|c\|_2}$, we have

$$\frac{\partial k}{\partial c} = \frac{\partial v}{\partial c} \cdot \frac{\partial k}{\partial v} \tag{12a}$$

$$= \frac{\partial v}{\partial c} \cdot \mathcal{B}^\top \tag{12b}$$

$$= \left( \frac{\mathbf{I}}{\|c\|_2} + \frac{\partial [(c^\top c)^{-\frac{1}{2}}]}{\partial c} \cdot c^\top \right) \cdot \mathcal{B}^\top \tag{12c}$$

$$= \left( \frac{\mathbf{I}}{\|c\|_2} + \frac{\partial [(c^\top c)]}{\partial c} \cdot (-\frac{1}{2} \frac{1}{\|c\|_2^3}) \cdot c^\top \right) \cdot \mathcal{B}^\top \tag{12d}$$

$$= \left( \frac{\mathbf{I}}{\|c\|_2} + 2c \cdot (-\frac{1}{2} \frac{1}{\|c\|_2^3}) \cdot c^\top \right) \cdot \mathcal{B}^\top \tag{12e}$$

$$= \frac{1}{\|c\|_2} \left( \mathbf{I} - \frac{c}{\|c\|_2} \cdot \frac{c^\top}{\|c\|_2} \right) \cdot \mathcal{B}^\top, \tag{12f}$$

where $\mathbf{I} \in \mathbb{R}^{(R-1) \times (R-1)}$ is the identity matrix. Combining the Equation (10) and (21), we can obtain the desired gradients of the loss function w.r.t the training parameter $c_i^t$, given as

$$\boxed{\frac{\partial \ell}{\partial c_i^t} = -\frac{1}{\|c_i^t\|_2} \left( \mathbf{I} - \frac{c_i^t}{\|c_i^t\|_2} \cdot \frac{c_i^{t\top}}{\|c_i^t\|_2} \right) \cdot \mathcal{B}^\top \cdot \left( P_{inv}^\top V^\top + U_{(t-1)p}^\top \Lambda \bar{K}_i^{t\top} \right) e} \tag{13}$$

**Influence functions** $\phi$: According to diffusion equation (4), the dependency of $u_t$ on the influence function $\phi_i^t$ is given as

$$u_t \to -\bar{K}_i^t \cdot \phi_i^t(K_i^t \cdot u_{(t-1)p}). \tag{14}$$

Let us define an auxiliary variable $y \in \mathbb{R}^O$ as

$$y = K_i^t \cdot u_{(t-1)p}. \tag{15}$$

In our work, function $\phi_i^t$ is represented as

$$\phi_i^t(z) = \sum_{j=1}^{M} w_{ij}^t \varphi\left( \frac{|z - \mu_j|}{\gamma} \right), \tag{16}$$

Therefore, the column vector $\phi_i^t(y) \in \mathbb{R}^O$ can be reformulated via a matrix equation

$$\phi_i^t(y) = G(y) \cdot w_i^t,$$

where $w_i^t \in \mathbb{R}^M$ is the vectorized version of parameters $w_{ij}^t$, matrix $G(y) \in \mathbb{R}^{O \times M}$ is given as

$$\underbrace{\begin{bmatrix} \varphi(\frac{|y_1-\mu_1|}{\gamma}) & \varphi(\frac{|y_1-\mu_2|}{\gamma}) & \cdots & \varphi(\frac{|y_1-\mu_M|}{\gamma}) \\ \varphi(\frac{|y_2-\mu_1|}{\gamma}) & \varphi(\frac{|y_2-\mu_2|}{\gamma}) & \cdots & \varphi(\frac{|y_2-\mu_M|}{\gamma}) \\ \vdots & \vdots & \ddots & \vdots \\ \varphi(\frac{|y_O-\mu_1|}{\gamma}) & \varphi(\frac{|y_O-\mu_2|}{\gamma}) & \cdots & \varphi(\frac{|y_O-\mu_M|}{\gamma}) \end{bmatrix}}_{G(y)} \underbrace{\begin{bmatrix} w_{i1} \\ w_{i2} \\ \vdots \\ w_{iM} \end{bmatrix}}_{w_i} = \underbrace{\begin{bmatrix} \phi_i(y_1) \\ \phi_i(y_2) \\ \vdots \\ \phi_i(y_Q) \end{bmatrix}}_{\phi_i(y)}. \tag{17}$$

4

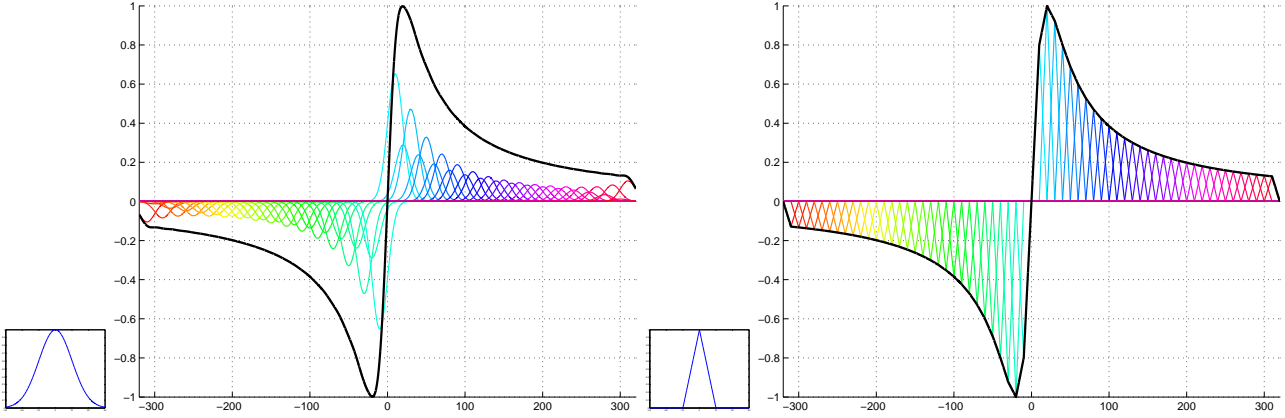
Figure 2. Function approximation via Gaussian $\varphi_g(z)$ or triangular-shaped $\varphi_t(z)$ radial basis function, respectively.

Now, it is straightforward to obtain $\frac{\partial u_t}{\partial w_i^t}$, given as

$$\frac{\partial u_t}{\partial w_i^t} = -G^\top \bar{K}_i^t{}^\top. \tag{18}$$

Then we can obtain the desired gradients of the loss function w.r.t the parameters of the influence function, written as

$$\boxed{\frac{\partial \ell}{\partial w_i^t} = -G^\top \bar{K}_i^t{}^\top e.} \tag{19}$$

In this paper, we investigate two typical RBFs [8]: (1) Gaussian radial basis $\varphi_g$ and (2) triangular-shaped radial basis $\varphi_t$, which are given as

$$\varphi_g(z) = \varphi\left(\frac{|z-\mu|}{\gamma}\right) = \exp\left(-\frac{(z-\mu)^2}{2\gamma^2}\right)$$

and

$$\varphi_t(z) = \varphi\left(\frac{|z-\mu|}{\gamma}\right) = \begin{cases} 1 - \frac{|z-\mu|}{\gamma} & |z-\mu| \leq \gamma \\ 0 & |z-\mu| > \gamma \end{cases}$$

respectively. The basis functions are shown in Figure 2, together with an example of the function approximation by using two different RBF methods.

In Figure 2, we can see that in the case of triangular-shaped RBF based function approximation any input variable $z$ only involves two basis functions, *i.e.*, each row of the $G$ matrix (17) only has two non-zero numbers. **Therefore, we can explicitly make use of this property in the implementation to speed up the computation of Equation (19), *i.e.*, the triangular-shaped RBF based training process is generally faster than the Gaussian RBF based one.**

In the training, the first order derivative of the influence function $\phi$ is also required as in Equation (13). In the case of Gaussian RBF, the first order derivative is given as

$$\phi_i'(z) = -\gamma \sum_{i=1}^M w_{ij} \exp\left(-\frac{\gamma}{2}(z-\mu_j)^2\right) \cdot (z-\mu_j).$$

In the case of triangular-shaped RBF, $\phi'$ is defined by a piece-wise constant function as $\phi$ is a piece-wise linear function. Although the influence function $\phi$ and its derivative $\phi'$ is not smooth, the training still works quite well.

In practice, in order to speed up the computation of the function value $\phi(z)$ and its derivative $\phi'(z)$ for the case of Gaussian RBF, we approximate these functions with piece-wise linear functions (the function values at discrete points are precomputed and stored in a lookup-table), then the function values at point $z$ can be retrieved efficiently using linear interpolation.

Concerning the function approximation accuracy, in general, the Gaussian RBF can provide a better approximation for generally smooth function than the triangular-shaped RBF with the same number of basis functions, as the latter generates a piece-wise linear function for approximation. In order to improve the approximation accuracy of the triangular-shaped RBF based method, usually we need to exploit more basis functions relative to the Gaussian RBF based method. However, using

more basis functions will bring an undesired problem of over-fitting. Therefore, certain regularization technique is required. Unfortunately, up to now we have not figured out the best choice for the regularization term.

In our work, we have investigated both function approximation methods, and we find that they generate similar results. We only present the results obtained by the Gaussian RBF due to space limitation. We do not provide a comprehensive comparison of two methods, as it is out of the scope of this paper.

### 2.2. Joint training

In the joint training, the parameters of all stages $T$ are optimized simultaneously. The joint training task is formulated as

$$\mathcal{L}(\Theta_{1,\cdots,T}) = \sum_{s=1}^{S} \ell(u_T^{(s)}, u_{gt}^{(s)}), \tag{20}$$

where the loss function only depends on $u_T$, the output of the final stage $T$. The gradients of the loss function w.r.t $\Theta_t$ is given as

$$\frac{\partial \ell(u_T, u_{gt})}{\partial \Theta_t} = \frac{\partial u_t}{\partial \Theta_t} \cdot \frac{\partial \ell(u_T, u_{gt})}{\partial u_t},$$

where $\frac{\partial u_t}{\partial \Theta_t}$ has been already done in the preceding subsection. Now the main issue is to calculate $\frac{\partial \ell(u_T, u_{gt})}{\partial u_t}$. As we only know

$$e = \frac{\partial \ell(u_T, u_{gt})}{\partial u_T} = T^\top (Tu_T - u_{gt}),$$

the standard back-propagation technique widely used in neural networks learning [10] can be used to calculate the desired gradients, which is written as

$$\frac{\partial \ell(u_T, u_{gt})}{\partial u_t} = \frac{\partial u_{t+1}}{\partial u_t} \cdot \frac{\ell(u_T, u_{gt})}{\partial u_{t+1}} \tag{21a}$$

$$= \frac{\partial u_{t+1}}{\partial u_t} \cdot \frac{\partial u_{t+2}}{\partial u_{t+1}} \cdot \frac{\ell(u_T, u_{gt})}{\partial u_{t+2}} \tag{21b}$$

$$= \frac{\partial u_{t+1}}{\partial u_t} \cdot \frac{\partial u_{t+2}}{\partial u_{t+1}} \cdots \frac{\partial u_T}{\partial u_{T-1}} \cdot e. \tag{21c}$$

In practice, (21) is computed using a backward manner starting from the last stage. Now the only thing we need to calculate is $\frac{\partial u_{t+1}}{\partial u_t}$. Recall the diffusion process shown in Figure 1, it is straightforward to see that

$$\frac{\partial u_{t+1}}{\partial u_t} = \frac{\partial u_{tp}}{\partial u_t} \cdot \frac{\partial u_{t+1}}{\partial u_{tp}},$$

where $\frac{\partial u_{tp}}{\partial u_t} = P_T^\top$ according to the equation $u_{tp} = P_T u_t$, and $\frac{\partial u_{t+1}}{\partial u_{tp}}$ can be obtained from the diffusion equation (4).

$$\frac{\partial u_{t+1}}{\partial u_{tp}} = (1-\lambda^{t+1})\mathbf{I} - \sum_{i=1}^{N_k} {K_i^{t+1}}^\top \cdot \Lambda_i \cdot \left(\bar{K}_i^{t+1}\right)^\top,$$

where $\Lambda_i$ is a diagonal matrix $\Lambda_i = \text{diag}(\phi_i^{t+1'}(z_1), \cdots, \phi_i^{t+1'}(z_p))$ with $z = k_i^{t+1} * u_{tp}$. Therefore, the overall $\frac{\partial u_{t+1}}{\partial u_t}$ is given as

$$\frac{\partial u_{t+1}}{\partial u_t} = P_T^\top \cdot \left((1-\lambda^{t+1})\mathbf{I} - \sum_{i=1}^{N_k} {K_i^{t+1}}^\top \cdot \Lambda_i \cdot \left(\bar{K}_i^{t+1}\right)^\top\right).$$

Then the gradients of $\frac{\partial \ell(u_T, u_{gt})}{\partial u_t}$ can be computed using the backward recurrence described above. Once we have obtained the results of $\frac{\partial \ell(u_T, u_{gt})}{\partial u_t}$, it is straightforward to calculate $\frac{\partial \ell(u_T, u_{gt})}{\partial \Theta_t}$ using the derivations in previous subsection.

## 3. Training for JPEG deblocking

As mentioned in the main paper, in this paper, we consider the JPEG deblocking problem by defining a new variational model, which incorporates the FoE image prior model and the quantization constraint set (QCS).

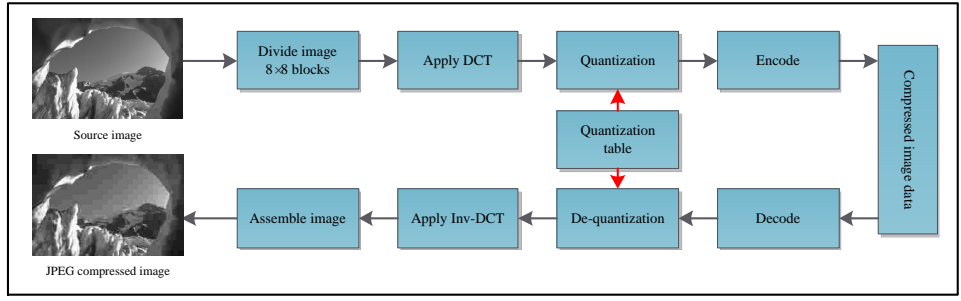

Figure 3. Schematic overview of the JPEG compression and decompression procedure

### 3.1. JPEG compression and the QCS

Figure 3 illustrates all the steps of the JPEG compression and decompression procedure. In the step of quantization, the transformed DCT coefficients of each $8 \times 8$ block are point-wise divided by the quantization matrix, and then the values are rounded to integer, which is where the loss of data takes place, as the rounding operation is a mapping of "$\infty \to 1$". Given an integer number $d$, any number in the interval $[d - \frac{1}{2}, d + \frac{1}{2}]$ is a possible candidate for the original number which is rounded to $d$.

With the compressed image data, we only know the integer coefficient data $(d^I_{i,j})_{1 \leq i,j \leq 8}$, where $I$ indicates a $8 \times 8$ block indexed by $I$, and the quantization matrix $(Q_{i,j})_{1 \leq i,j \leq 8}$. Therefore, the possible original DCT coefficients, which yield $(d^I_{i,j})$ during the quantization and rounding procedure are given by the interval

$$S^I_{i,j} = [Q_{i,j}(d^I_{i,j} - \frac{1}{2}), Q_{i,j}(d^I_{i,j} + \frac{1}{2})].$$

This result is for the block $I$. For the full size image, we just need to repeat this result for each distinct block. All the intervals $S^I_{i,j}$ associated with each $8 \times 8$ block form the so-called QCS, which is simply a box constraint determining all possible source images.

### 3.2. Variational model for image deblocking

In our training, an image $u$ of size $m \times n$ is padded with $\omega$ pixels (we set $\omega = 8$ for this problem). In order to simplify the notation, the interval $S$ is represented by two column vectors $a \in \mathbb{R}^O$ and $b \in \mathbb{R}^O$ ($O = (m + 2\omega) \times (n + 2\omega)$), which correspond to the lower and upper bounds of the intervals $S^I_{i,j}$, respectively. We further define a highly sparse matrix $D \in \mathbb{R}^{O \times O}$, which makes $Du$ equivalent to the block-wise DCT transform applied to the two-dimensional image $u$.

Given the compressed image data, the QCS is given as the box constraint $S = [a, b]$, and the set of possible source image of the compression process is defined as

$$U = \{u \in \mathbb{R}^O \mid (Du)_p \in [a_p, b_p]\}.$$

Then we can define our variational model based on the FoE image prior model and QCS, which reads as

$$\arg\min_{u \in U} E(u) = \sum_{i=1}^{N_k} \rho_i(k_i * u). \tag{22}$$

This is a constrained optimization problem, and it can be rewritten as

$$\arg\min_u E(u) = \sum_{i=1}^{N_k} \rho_i(k_i * u) + \mathcal{I}_S(Du), \tag{23}$$

where

$$\mathcal{I}_S(Du) = \begin{cases} 0 & \text{if } Du \in S, \\ \infty & \text{else}. \end{cases}$$

In this formulation, we exploit the convex set $S$ instead of set $U$, as $S$ is a box constraint, which is simpler than $U$.

7

As the minimization problem (23) contains the non-smooth indicator function, the standard gradient descent algorithm is not applicable. Therefore, we resort to the more general proximal gradient method [11], which is applicable to solve a class of the following minimization problems

$$\min_u h(u) = f(u) + g(u), \tag{24}$$

where $f$ is a smooth function and $g$ is convex (possibly non-smooth) function. The proximal gradient method is defined as

$$u_t = (\mathbf{I} + \tau \partial g)^{-1}(u_{t-1} - \tau \nabla f(u_{t-1})),$$

where $\tau$ is the step size parameter, $(\mathbf{I} + \tau \partial g)^{-1}$ denotes the proximal mapping operator. Casting the problem (23) in the form of (24), we have $f(u) = \sum_{i=1}^{N_k} \rho_i(k_i * u)$ and $g(u) = \mathcal{I}_S(Du)$. It is easy to check that

$$\nabla f(u) = \sum_{i=1}^{N_k} \bar{k}_i * \phi_i(k_i * u),$$

with $\phi_i(z) = \rho_i'(z)$. We again make the modification $\bar{k}_i$ to the rigorous formulation $K_i^\top$. The proximal mapping with respect to $g$ is given as the following minimization problem

$$(I + \tau \partial g)^{-1}(\hat{u}) = \arg\min_u \frac{\|u - \hat{u}\|_2^2}{2} + \tau \mathcal{I}_S(Du). \tag{25}$$

As DCT is a orthogonal transform, i.e., $D^\top D = DD^\top = \mathbf{I}$, then we have

$$\|Du - D\hat{u}\|_2^2 = (u - \hat{u})^\top D^\top D(u - \hat{u}) = (u - \hat{u})^\top (u - \hat{u}) = \|u - \hat{u}\|_2^2.$$

For problem (25), let

$$c = Du, \ \hat{c} = D\hat{u}.$$

Note that the connection between $c$ and $u$ (also $\hat{c}$ and $\hat{u}$) is a mapping of one-to-one.

It turns out that

$$\arg\min_u \frac{\|u - \hat{u}\|_2^2}{2} + \tau \mathcal{I}_S(Du) \iff \arg\min_c \frac{\|c - \hat{c}\|_2^2}{2} + \tau \mathcal{I}_S(c).$$

Obviously, the solution for the minimization problem of right side is given as the following point-wise projection onto the interval, *i.e.*, QCS

$$\tilde{c}_p = \begin{cases} \hat{c}_p & \text{if } \hat{c}_p \in S_p = [a_p, b_p] \\ b_p & \text{if } \hat{c}_p > b_p \\ a_p & \text{if } \hat{c}_p < a_p. \end{cases} \tag{26}$$

Finally, the the solution of $u$ is given as $\tilde{u} = D^\top \tilde{c}$. Therefore, the overall gradient descent step is given as

$$u_t = D^\top \text{proj}_{\text{QCS}} \left( D \left( u_{t-1} - \sum_{i=1}^{N_k} \bar{k}_i * \phi_i(k_i * u_{t-1}) \right) \right), \tag{27}$$

where $\text{proj}_{\text{QCS}}(\cdot)$ denotes the orthogonal projection onto QCS (26). In our training model, as we consider different filters and influence functions for each stage $t$, the accurate diffusion process is given as

$$u_t = D^\top \text{proj}_{\text{QCS}} \left( D \left( u_{t-1} - \sum_{i=1}^{N_k} \bar{k}_i^t * \phi_i^t(k_i^t * u_{t-1}) \right) \right). \tag{28}$$

The projection operator can be represented by the function $\eta(z)$ show in Figure 4. Therefore, the corresponding gradient descent step (28) can be rewritten as the following formulas by introducing the auxiliary variables $v, z,$ and $y$.

$$\begin{cases} u_t = D^\top v \\ v = \eta(z) \\ z = Dy \\ y = u_{t-1} - \sum_{i=1}^{N_k} \bar{k}_i^t * \phi_i^t(k_i^t * u_{t-1}). \end{cases} \tag{29}$$

8

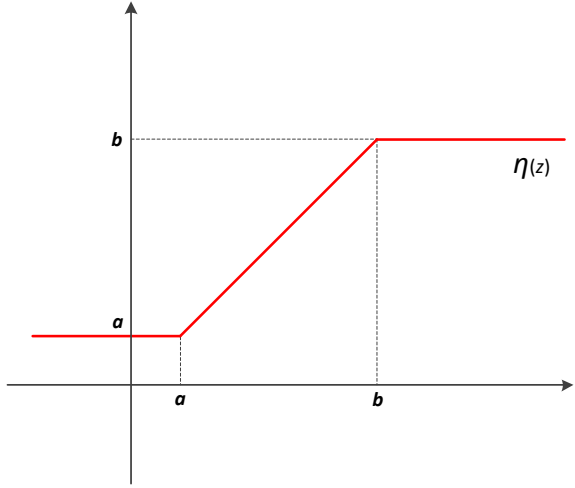

Figure 4. The projection function $\eta(z)$.

For the training of this new problem, the training parameters $\Theta_t$ is given by $\Theta_t = \{\phi_i^t, k_i^t\}$. According to the result of (5), for this new training problem, we can make use of the framework presented in the last section and we just need to recalculate $\frac{\partial u_t}{\partial \Theta_t}$ from (29), which is given as

$$\begin{aligned}\frac{\partial u_t}{\partial \Theta_t} &= \frac{\partial y}{\partial \Theta_t} \cdot \frac{\partial z}{\partial y} \cdot \frac{\partial v}{\partial z} \cdot \frac{\partial u_t}{\partial v} \\ &= \frac{\partial y}{\partial \Theta_t} \cdot D^\top \cdot \mathrm{diag}(\eta'(z)) \cdot D \,,\end{aligned} \quad (30)$$

where $\eta'(z) = (\eta'(z_1), \eta'(z_2), \cdots, \eta'(z_O))^\top \in \mathbb{R}^O$ with $\eta'(z_p)$ defined as

$$\eta'(z_p) = \begin{cases} 1 & \text{if } z_p \in [a_p, b_p], \\ 0 & \text{else} \end{cases} \quad (31)$$

Even though we do not consider any smoothing technique for the non-smooth function $\eta$, in practice we find that it is not a problem for the training procedure by using the above discontinuous derivative. According to the derivations (9) and (18) in the last section, it is straightforward to calculate $\frac{\partial y}{\partial \Theta_t}$, which leads to exactly the same results.

Concerning the joint training model, we can still make use of the same framework presented in the last section and we just need to additionally compute $\frac{\partial u_t}{\partial u_{t-1}}$. Taking into account the operation of boundary handling, $\frac{\partial u_t}{\partial u_{t-1}}$ is given as

$$\begin{aligned}\frac{\partial u_t}{\partial u_{t-1}} &= P_T^\top \cdot \frac{\partial y}{\partial u_{t-1}} \cdot \frac{\partial z}{\partial y} \cdot \frac{\partial v}{\partial z} \cdot \frac{\partial u_t}{\partial v} \\ &= P_T^\top \cdot \left( \mathbf{I} - \sum_{i=1}^{N_k} K_i^{t\top} \cdot \Lambda_i \cdot \bar{K}_i^{t\top} \right) \cdot D^\top \cdot \mathrm{diag}(\eta'(z)) \cdot D\end{aligned} \quad (32)$$

Combining the derivation results of (30), (32) and the framework presented in the last section, we can reach the formulas required for the training of the JPEG deblocking problem.

## 4. Denoising and deblocking examples

In this section, we provide examples to illustrate the performance of our trained nonlinear diffusion processes for Gaussian denoising and JPEG deblocking. See Figure 5 and Figure 6 for image denoising examples for noise level $\sigma = 25$ on the images from the test dataset. Note the differences in the highlighted regions.

We also present the corresponding runtime either on CPU or GPU. We do not count the memory transfer time between CPU/GPU for both GPU implementations (if counted, the run time will nearly double). Note the speed gap between $\mathrm{CSF}_{7 \times 7}^5$

9

and our model is not mainly caused by the different Matlab/CUDA implementations, but the lower computational expense of our approach.

Figure 7 presents a denoising example on a megapixel-size natural image of size $1050 \times 1680$. Overall, nonlocal methods (BM3D and WNNM) are prone to generating artifacts. The $\text{TRD}_{7\times 7}^5$ model provides the highest PSNR value, and better preserves tiny image structures, such as the tree branches shown in the highlighted region. Furthermore, our method exhibits the best runtime.

Figure 8 presents four JPEG deblocking examples for the compression quality $q = 10$. Note the effectiveness of our trained TRD model, meanwhile, remember that our approach is extremely fast on GPUs.

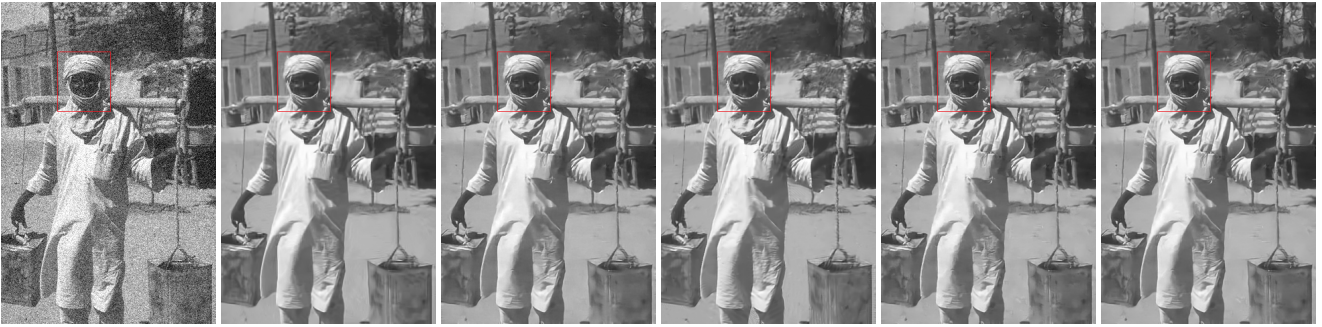

(a) Noisy, 20.17dB    (b) BM3D, 28.60dB    (c) $\text{CSF}^5_{7\times7}$, 28.83dB    (d) WNNM, 28.82dB    (e) $\text{TRD}^5_{5\times5}$, 28.85dB    (f) $\text{TRD}^5_{7\times7}$, 28.97dB

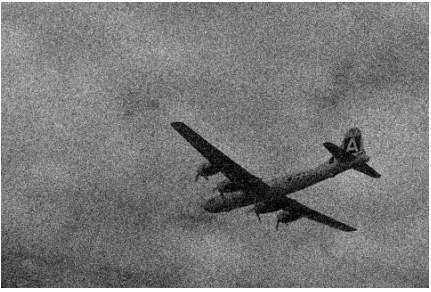 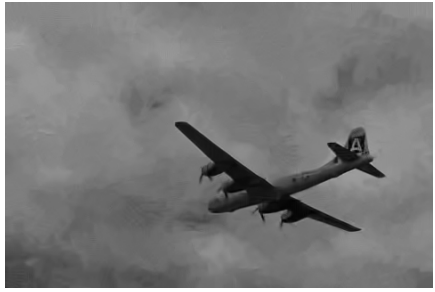 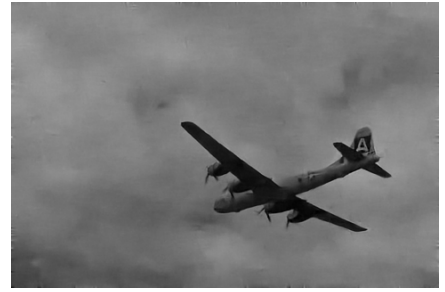

(g) Noisy, 20.17dB    (h) BM3D, 36.78dB/**CPU: 2.5s**    (i) $\text{CSF}^5_{7\times7}$, 37.15dB/**GPU: 0.55s**

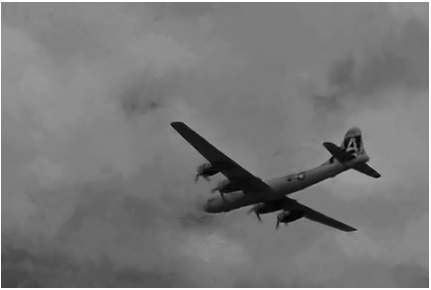 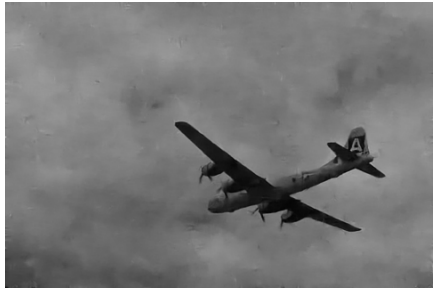 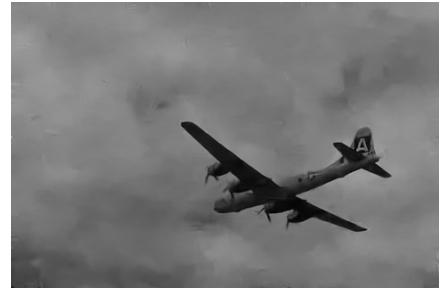

(j) WNNM, 36.95dB/**CPU: 393.2s**    (k) $\text{TRD}^5_{5\times5}$, 37.04dB/**GPU: 9.1ms**    (l) $\text{TRD}^5_{7\times7}$, 37.64dB/**GPU: 20.3ms**

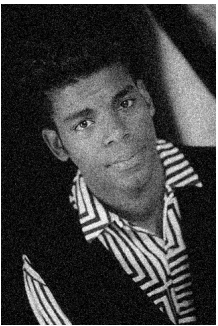 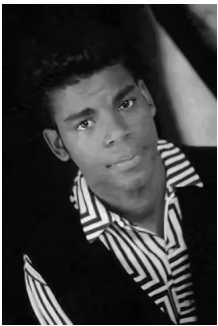 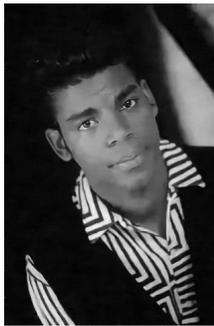 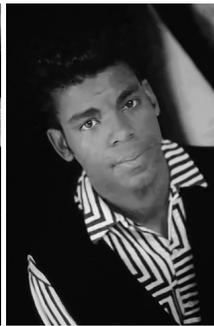 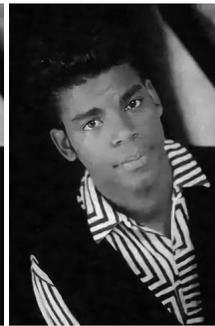 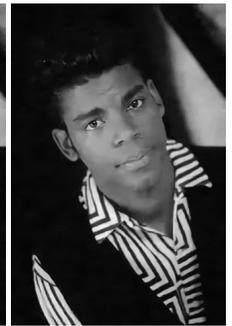

(m) Noisy, 20.17dB    (n) BM3D, 33.24dB    (o) $\text{CSF}^5_{7\times7}$, 32.93dB    (p) WNNM, 33.77dB    (q) $\text{TRD}^5_{5\times5}$, 32.91dB    (r) $\text{TRD}^5_{7\times7}$, 33.34dB

Figure 5. Denoising results on three test images ($\sigma = 25$) by different methods (compared with BM3D [5], WNNM [7] and CSF model [12]), together with the corresponding computation time either on CPU or GPU. Note that for the last image, which contains many repeated local patterns, *e.g.*, the T-shirt region, the nonlocal methods (BM3D and WNNM) generally should work better than the local methods (CSF model and our TRD model), because the nonlocal models explicitly exploit nonlocal self-similarity across the image. Even though the nonlocal methods can benefit from these repeated local patterns, our trained $\text{TRD}^5_{7\times7}$ still show strongly competitive performance. **Best viewed magnified on screen. Note the differences in the highlighted region.**

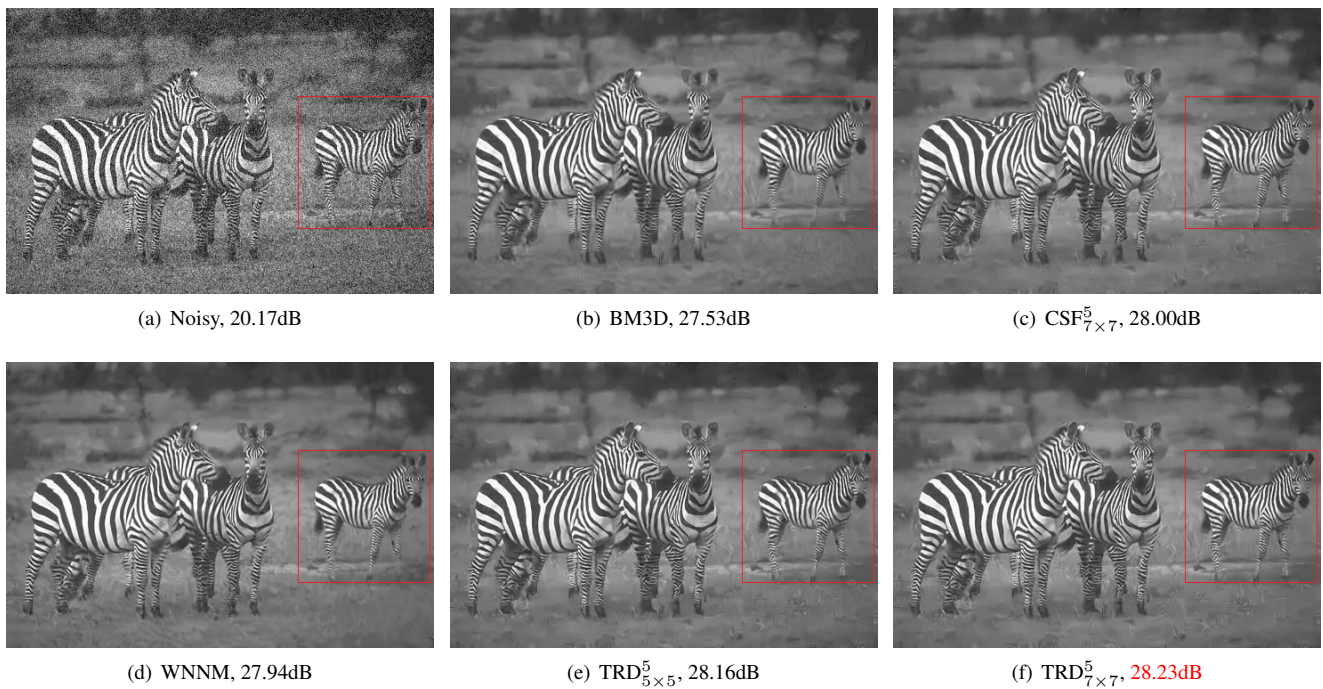

Figure 6. Another example noise level $\sigma = 25$. Note the differences in the highlighted region.

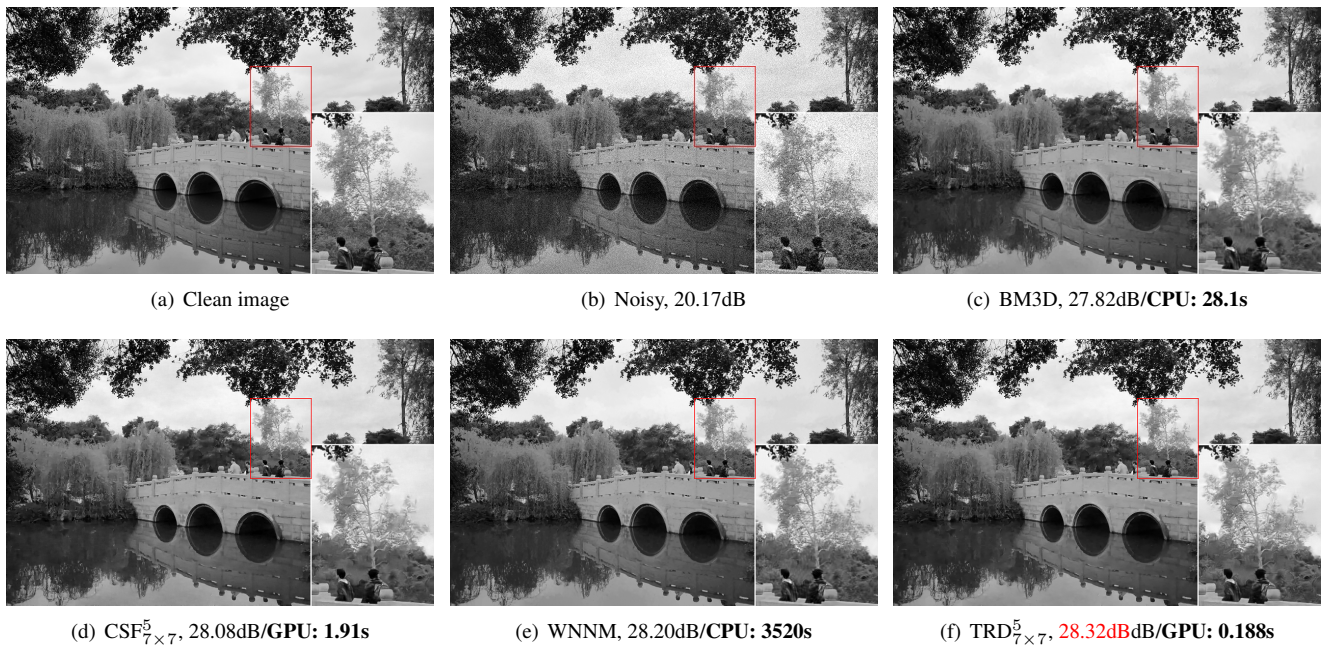

Figure 7. Denoising example on a high resolution "Chinese bridge" image of size $1050 \times 1680$ ($\sim 1.68$ mega-pixels) for noise level $\sigma = 25$. Compared with BM3D [5], WNNM [7] and CSF model [12], together with the corresponding runtime. Our $\text{TRD}^5_{7\times 7}$ model achieves the highest PSNR value, and it better preserves tiny image structures, *e.g.*, the tree branches in the highlighted region. (**Best viewed magnified on screen.**) Moreover, our model offers remarkably preferable runtime performance based on GPU implementation.

12

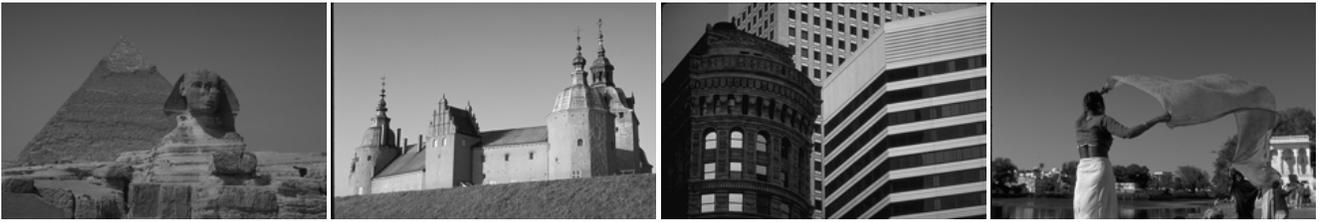

Fig-1 Clean image | Fig-2 Clean image | Fig-3 Clean image | Fig-4 Clean image

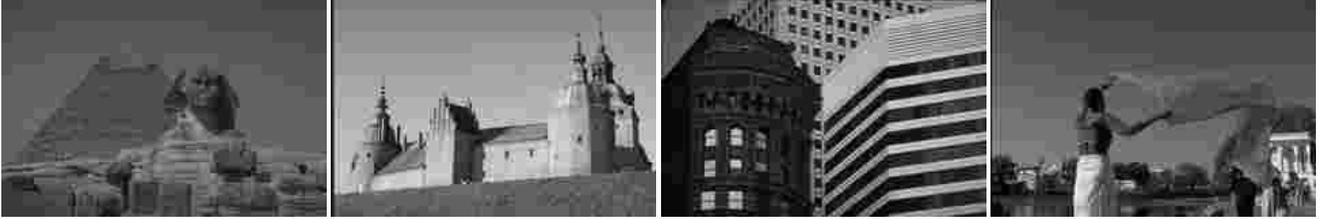

Fig-5 Lossy image (30.02) | Fig-6 Lossy image (27.59) | Fig-7 Lossy image (24.24) | Fig-8 Lossy image (28.56)

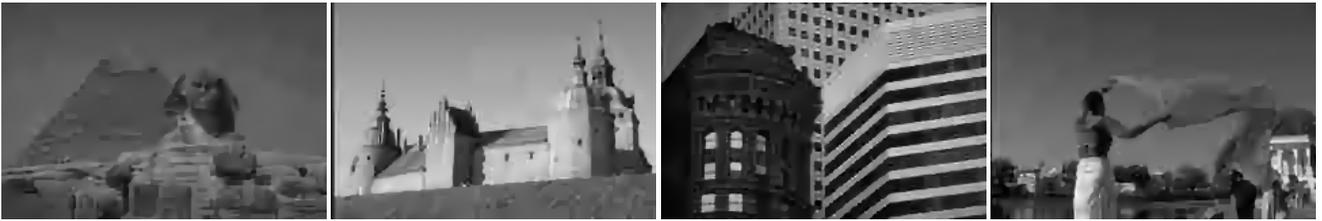

Fig-9 TGV[2] (30.44) | Fig-10 TGV[2] (28.31) | Fig-11 TGV[2] (25.12) | Fig-12 TGV[2] (29.34)

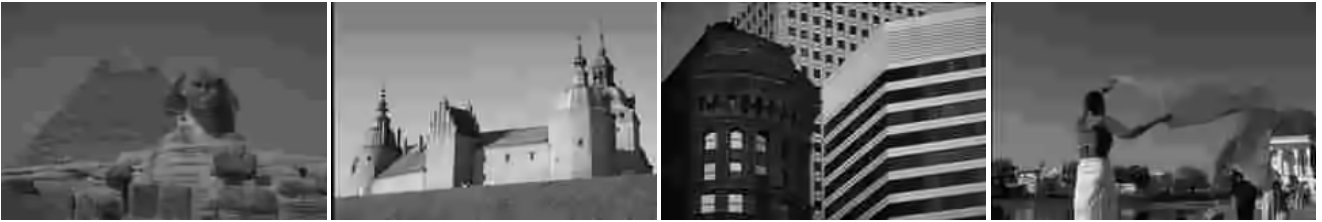

Fig-13 Dic. SR[3] (30.50) | Fig-14 Dic. SR[3] (28.20) | Fig-15 Dic. SR[3] (25.11) | Fig-16 Dic. SR[3] (29.29)

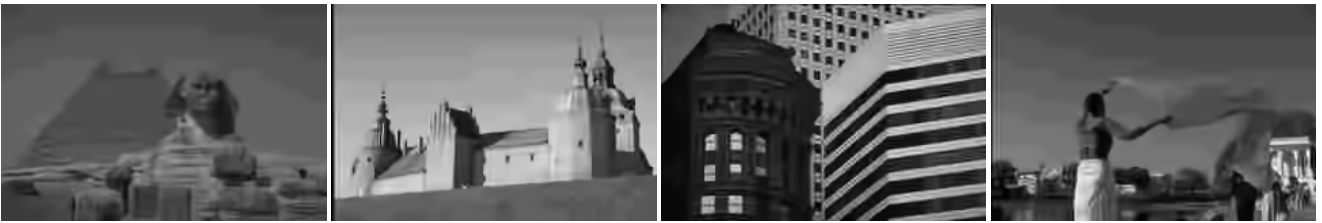

Fig-17 SADCT[6] (30.69) | Fig-18 SADCT[6] (29.63) | Fig-19 SADCT[6] (25.54) | Fig-20 SADCT[6] (29.53)

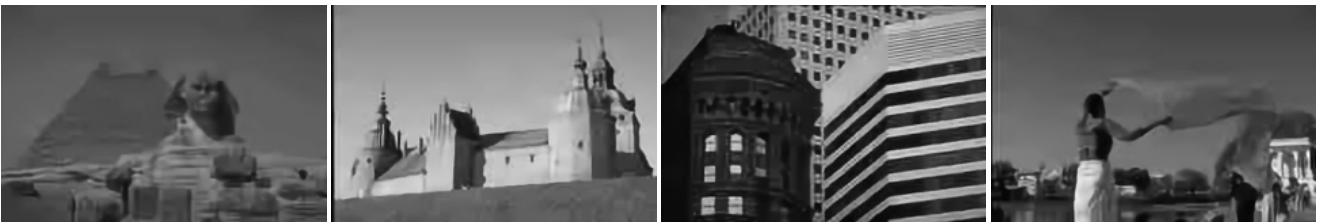

Fig-21 RTF[9] (30.93) | Fig-22 RTF[9] (29.16) | Fig-23 RTF[9] (26.37) | Fig-24 RTF[9] (29.94)

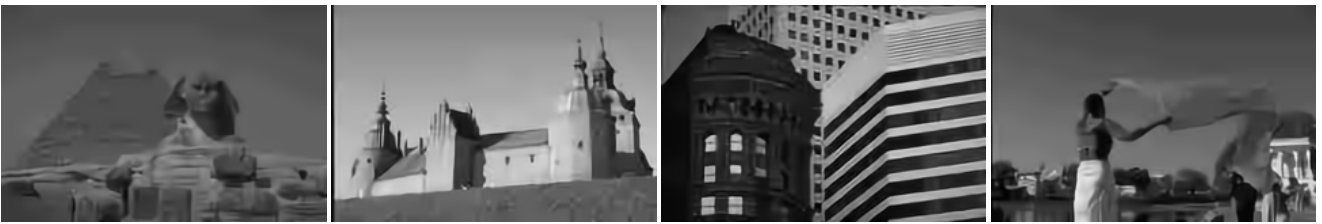

Fig-25 $\text{TRD}_{7\times 7}^4$ (31.04) | Fig-26 $\text{TRD}_{7\times 7}^4$ (29.47) | Fig-27 $\text{TRD}_{7\times 7}^4$ (27.05) | Fig-28 $\text{TRD}_{7\times 7}^4$ (30.17)

Figure 8. Image deblocking for images compressed by JPEG encoder with the quality $q = 10$. **Note the differences in the sky.**



\end{document}